\documentclass[a4paper]{article}
\usepackage{arxiv}

\usepackage[utf8]{inputenc} 
\usepackage[T1]{fontenc}    
\usepackage{nicefrac}       
\usepackage{microtype}      

\usepackage{cite}
\usepackage{amsmath,amssymb,amsfonts}
\usepackage{algorithmic}
\usepackage{graphicx}
\usepackage{rotating}
\usepackage{pdflscape}
\usepackage{subcaption}
\usepackage{textcomp}
\usepackage{multicol}
\usepackage{mparhack}

\usepackage[inkscapeformat=png]{svg}
\definecolor{linkcolor}{RGB}{120, 45, 17}
\usepackage{xspace} 
\def\BibTeX{{\rm B\kern-.05em{\sc i\kern-.025em b}\kern-.08em
    T\kern-.1667em\lower.7ex\hbox{E}\kern-.125emX}}

\makeatletter
\def\els@aparagraph[#1]#2{\elsparagraph[#1]{#2\@addpunct{.}}}
\def\els@bparagraph#1{\elsparagraph*{#1\@addpunct{.}}}
\makeatother

\usepackage{flushend} 
\usepackage{url}
\usepackage[colorlinks=true,allcolors=linkcolor]{hyperref}
\usepackage{colortbl}
\usepackage{lineno}
\usepackage[noabbrev]{cleveref}
\usepackage[finalnew]{trackchanges}
\usepackage{amssymb}
\usepackage{pifont}
\newcommand{\xmark}{\ding{55}}%

\date{}

\title{A Comprehensive Survey of Deep Transfer Learning for Anomaly Detection in Industrial Time Series: Methods, Applications, and Directions}

\renewcommand{\headeright}{}
\renewcommand{\undertitle}{}

\author{\href{https://orcid.org/0009-0006-0236-4707}{\includegraphics[scale=0.06]{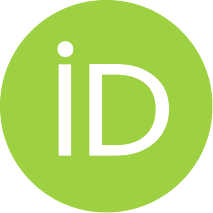}\hspace{1mm}Peng Yan} \\
Centre for Artificial Intelligence\\
ZHAW School of Engineering\\
Winterthur, ZH, Switzerland\\
	\texttt{yanp@zhaw.ch} \\
\And
\href{https://orcid.org/0000-0003-4679-8081}{\includegraphics[scale=0.06]{pic/orcid.pdf}\hspace{1mm}Ahmed Abdulkadir} \\
Centre for Artificial Intelligence\\
ZHAW School of Engineering\\
Winterthur, ZH, Switzerland\\
	\texttt{abdk@zhaw.ch} \\
\And
\href{https://orcid.org/0009-0007-0851-665X}{\includegraphics[scale=0.06]
{pic/orcid.pdf}\hspace{1mm}Paul-Philipp Luley} \\
Centre for Artificial Intelligence\\
ZHAW School of Engineering\\
Winterthur, ZH, Switzerland\\
	\texttt{lule@zhaw.ch} \\
\And
\href{https://orcid.org/0000-0002-7577-783X}{\includegraphics[scale=0.06]
{pic/orcid.pdf}\hspace{1mm}Matthias Rosenthal} \\
Institute of Embedded Systems\\
ZHAW School of Engineering\\
Winterthur, ZH, Switzerland\\
	\texttt{rosn@zhaw.ch} \\
\And
\href{https://orcid.org/0009-0002-5760-9346}{\includegraphics[scale=0.06]
{pic/orcid.pdf}\hspace{1mm}Gerrit A. Schatte} \\
Innovation Lab\\
Kistler Instrumente AG\\
Winterthur, ZH, Switzerland \\
	\texttt{gerrit.schatte@kistler.com} \\
\And
\href{https://orcid.org/0000-0001-8560-2120}{\includegraphics[scale=0.06]{pic/orcid.pdf}\hspace{1mm}Benjamin F. Grewe} \\
Institute of Neuroinformatics\\
ETH \& University of Z\"urich\\
Z\"urich, ZH, Switzerland\\
	\texttt{benjamin.grewe@uzh.ch} \\
\And
\href{https://orcid.org/0000-0002-3784-0420}{\includegraphics[scale=0.06]{pic/orcid.pdf}\hspace{1mm}Thilo Stadelmann} \\
ZHAW Centre for Artificial Intelligence \& European Centre for Living Technology (ECLT)\\
Winterthur, ZH, Switzerland \& Venice, Veneto, Italy\\
	\texttt{stdm@zhaw.ch} }

\hypersetup{
pdftitle={A Comprehensive Survey of Deep Transfer Learning for Anomaly Detection in Industrial Time Series},
pdfsubject={}, 
pdfauthor={Peng Yan, Ahmed Abdulkadir, Paul-Philipp Luley, Matthias Rosenthal, Gerrit A. Schatte, Benjamin F. Grewe, and Thilo Stadelmann}, 
pdfkeywords={deep transfer learning, time series analysis, anomaly detection, manufacturing process monitoring, predictive maintenance }}

\usepackage{booktabs}

\begin{document}
\maketitle

\begin{abstract}
Automating the monitoring of industrial processes has the potential to enhance efficiency and optimize quality by promptly detecting abnormal events and thus facilitating timely interventions.
Deep learning, with its capacity to discern non-trivial patterns within large datasets, plays a pivotal role in this process.
Standard deep learning methods are suitable to solve a specific task given a specific
type of data.
During training, deep learning demands large volumes of labeled data.
However, due to the dynamic nature of the industrial processes and environment, it is impractical
to acquire large-scale labeled data for standard deep learning training for every slightly different case anew.
Deep transfer learning offers a solution to this problem.
By leveraging knowledge from related tasks and accounting for variations in data distributions, the transfer learning framework solves
new tasks with little or even no additional labeled data.
The approach bypasses the need to retrain a model from scratch for every new setup and dramatically reduces the labeled data requirement.
This survey first provides an in-depth review of deep transfer learning, examining the problem settings of transfer learning and classifying the prevailing deep transfer learning methods. Moreover, we delve into applications of deep transfer learning in the context of a broad spectrum of time series anomaly detection tasks prevalent in primary industrial domains, e.g., manufacturing process monitoring, predictive maintenance, energy management, and infrastructure facility monitoring.
We discuss the challenges and limitations of deep transfer learning in industrial contexts and conclude the survey with practical directions and actionable suggestions to address the need to leverage diverse time series data for anomaly detection in an increasingly dynamic production environment.


\end{abstract}

\clearpage
\section{Introduction}
\subsection{motivation and contribution}
The fourth industrial revolution -- Industry 4.0 \cite{kagermann2011industrie}, that is characterized by
increasing efficiency through the digitization of production, automation, and horizontal integration across companies \cite{Roblek2016},
and the advent of connected cyber-physical systems -- referred to as internet of things \cite{Wang2018, Jeschke2017, Dalenogare2018}, increases the need for autonomous and intelligent process monitoring.
This can be exemplified by the use case of a smart factory in which industrial processes are transformed to be more flexible, intelligent, and dynamic \cite{Kagermann2017},
or the use case of decentralized energy production with wind and solar \cite{9381850}.
In these examples, AI-powered anomaly detection integrates the analysis of time series data to detect unusual patterns in the recorded data.
To achieve this, a deep learning architecture is modeled to capture indicators of normal and abnormal operation.
The learning process involves the analysis of historic time series sensor data of normal and possibly abnormal operations.
This data is for example used for \textsl{representation-} or \textsl{reconstruction-based} learning.
After training, the deep learning model \textsl{represents} or \textsl{reconstructs} normal data in a certain way.
The model is designed in a way that abnormal data--because it is different--is either \textsl{represented} differently from the normal data or
\textsl{reconstructed} poorly and thus recognized as an anomaly.
By identifying operational parameters that fall outside a window of normal interval, operators can trigger interventions and adjustments to ensure high product quality and safe operations.
To achieve this, physical properties such as pressure or temperature are monitored and analyzed in real-time applications.
Changes in these variables capture drifting and abrupt faults caused by process failures or malfunctions \cite{parkReviewFaultDetection2020}.
The production process must adapt quickly to changes in production and the environment to meet the requirements for flexibility and dynamics. Further use cases exist in a wide range of diverse fields, such as manufacturing monitoring including automatic quality control \cite{liao2021manufacturing, lockner2021induced}, predictive maintenance of goods and services \cite{maschler2021towards, wen2019time, xu2019digital, mao2020robust, wang2021anomaly, canizoMultiheadCNNRNN2019}, infrastructure monitoring of building systems \cite{weberDetectionBuildingOccupancy2020, sayedTimeseries2DImages2023} and power plant \cite{yao2022model}, digital agriculture \cite{abdallah2021anomaly}, petrochemical process optimization \cite{panjapornpon2023explainable}, computer network intrusion detection \cite{dhillon2020towards}, or aircraft flight monitoring \cite{xiong2018application}, to name a few.

Artificial Intelligence, particularly deep learning, provides competent frameworks with underlying deep neural networks to automate intelligent monitoring and provide valuable assistance to operators and high-level control systems. 
Leveraging the power of deep learning, informative features of the data -- technically referred to as \textsl{representations} \cite{Bengio2014_Represen} -- can be captured in a machine-learned model and thereby enable a detailed understanding of variations in standard operations.

However, the task or underlying data may change under non-trivial and non-stationary conditions.
For instance, the monitoring system of a milling machine may be assigned the task of identifying a blunt tool based on vibration in one scenario, and in a different scenario, it may utilize the same vibration measurements to detect insufficient cooling lubricant.
Knowledge acquired to solve one task in one setting with a given tool, machined part, and type of machine may
be transferred to solve the same or similar task in another setting with a different tool, machined part, or type of machine.
Slowly changing conditions (drifts), abrupt mode changes (for instance, due to tool change), and new tasks (such as the detection of another failure mode) may require adjustments to the deep learning model.
In these cases, it is desirable to adjust the analysis model without retraining from scratch, as it is costly or impractical to acquire sufficient training data to learn the full manifold \cite{Maschler2022}.

\begin{figure}[t!]
\centering
\includegraphics[width=0.6\columnwidth]{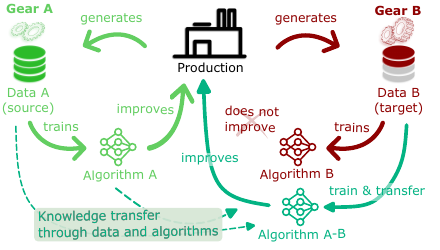}
\caption{\label{fig_intro}Transfer
learning is useful when changes in production take place and sufficient data
for full retraining is not available as shown here for a hypothetical production
of two types of gears.
In the production of gear A, a lot of data is available to train a deep
learning model that helps improve production.
In the production of gear B, data is more limited, and the traditionally trained deep learning model fails to improve production.
With suitable transfer learning methods, however, data and algorithms acquired during
the production of gear A can be leveraged to support improving the production
of gear B because the data and tasks in the production of both gears are related.}
\end{figure}

Transfer learning is a machine learning framework to achieve this \cite{Maschler2021TLinIA,zhuang2020comprehensive, pan2010survey, tan2018survey, yosinski2014transferable}.
As depicted in Fig. \ref{fig_intro}, data and algorithms from one task may be leveraged in a new related one.
By accounting for changes in data distributions and tasks or leveraging existing models, knowledge learned from related tasks can be used to improve performance on
new tasks instead of retraining a model for each individual application from scratch. 
This transfer-learning-boosted modeling forms the basis for identifying anomalies that deviate from established patterns in a non-trivial manner without full re-training.

Deep transfer learning \cite{tan2018survey,yuSurveyDeepTransfer2022} extends the transfer learning paradigm by leveraging deep learning. 
In industrial contexts, it ensures optimal production even as production conditions shift.
This dynamic adaptability is key in maintaining the effectiveness of anomaly detection systems in the dynamic environment that characterizes industrial applications
including the broad categories of manufacturing process monitoring, predictive maintenance, energy management, and infrastructure facility monitoring as detailed in Section \ref{sec_industrialapplications}.

This survey is a non-systematic yet application-oriented review with a narrow focus on deep transfer learning for anomaly detection in time series in the industry. Our main contributions are as follows: 
\begin{itemize}
    \item We categorize transfer learning problem settings and then systematically summarize deep transfer learning approaches into four categories. With the foundations of deep transfer learning, we equip the reader with a working knowledge of the main principles and intuitions.
    \item We analyze the recent literature and provide a comprehensive overview of the current state of the art of deep transfer learning approaches for time series anomaly detection for main industrial applications.
    \item We discuss potential challenges and limitations and then give directions for future work with actionable recommendations for AI practitioners and decision-makers.
\end{itemize}

To our knowledge, this is the first survey of deep transfer learning in the narrow context of industrial time series anomaly detection.
The review describes the underlying methodological principles and methods within a generic taxonomy and discusses practical implications for AI practitioners to make informed decisions.
We cover multiple areas of application, including manufacturing monitoring, maintenance prediction, and infrastructure monitoring.

The rest of the paper is organized as follows. First, we provide an overview of transfer learning by introducing a taxonomy of transfer learning problem settings and further categorizing deep transfer learning approaches (Section \ref{sec_tl}). Then, we describe the task of anomaly detection in time series (Section \ref{sec_tsanomalydetection}) in selected industrial applications (Section \ref{sec_industrialapplications}). To conclude, we discuss current challenges, limitations, and future research directions (Sections \ref{sec_discussion}--\ref{sec_conclusions}) in the field.

\subsection{Survey methodology}
\label{sec_methodology}
\begin{figure}[t!]
\centering
\includegraphics[width=0.6\columnwidth]{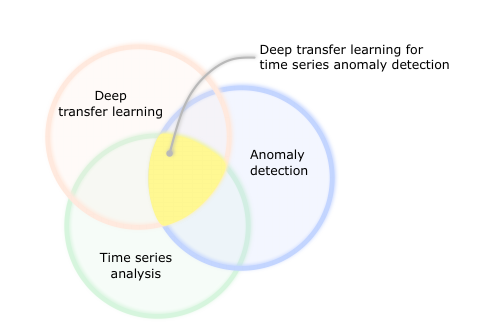}
\caption{\label{survey_topic} Venn diagram of this survey's focus on the intersection of transfer learning, anomaly detection, and time series analysis.}
\end{figure}

We seek to identify application-oriented peer-reviewed literature in the intersection of transfer learning as the learning framework, time series as the data domain, and anomaly detection as the task (Fig. \ref{survey_topic}).
To execute the selection process of literature, we search related terms on Google Scholar, Scopus, Elsevier, and IEEE databases.
Based on the title, we pick those papers that may fit the narrow topic into a pre-selection list.
Eventually, we included publications matching the topic according to the abstract and screening of the content.

Along the reviewed topical papers, we include contextually relevant papers such as deep learning approaches that are agnostic to data types and tasks. For deep transfer learning in general, we searched the keywords ``transfer learning'' and ``deep transfer learning''. Specifically, we focus more on deep transfer learning approaches. 
Then, we switch to the application-oriented cases where deep transfer learning is applied to tackle time series anomaly detection in the main industrial applications. To achieve this, we search queries like ``deep transfer learning for time series anomaly detection'' and ``deep transfer learning for predictive maintenance''.  
After searching in the database, we carefully check and screen out the most relevant literature based on the following inclusion/exclusion criteria: (1) We only include the applications that utilize deep transfer learning approaches, instead of traditional transfer learning;
(2) We only include publications after 2013;
(3) We cover all three main topics in Fig. \ref{survey_topic} (highlighted with cycles), but we specifically focus on the intersection of the three aforementioned domains.
After carefully screening out, we select 45 papers for deep transfer learning in general and 37 papers for deep transfer learning for anomaly detection in industrial time series.

\begin{figure*}[ht!]
\centering
\includegraphics[width=1\textwidth]{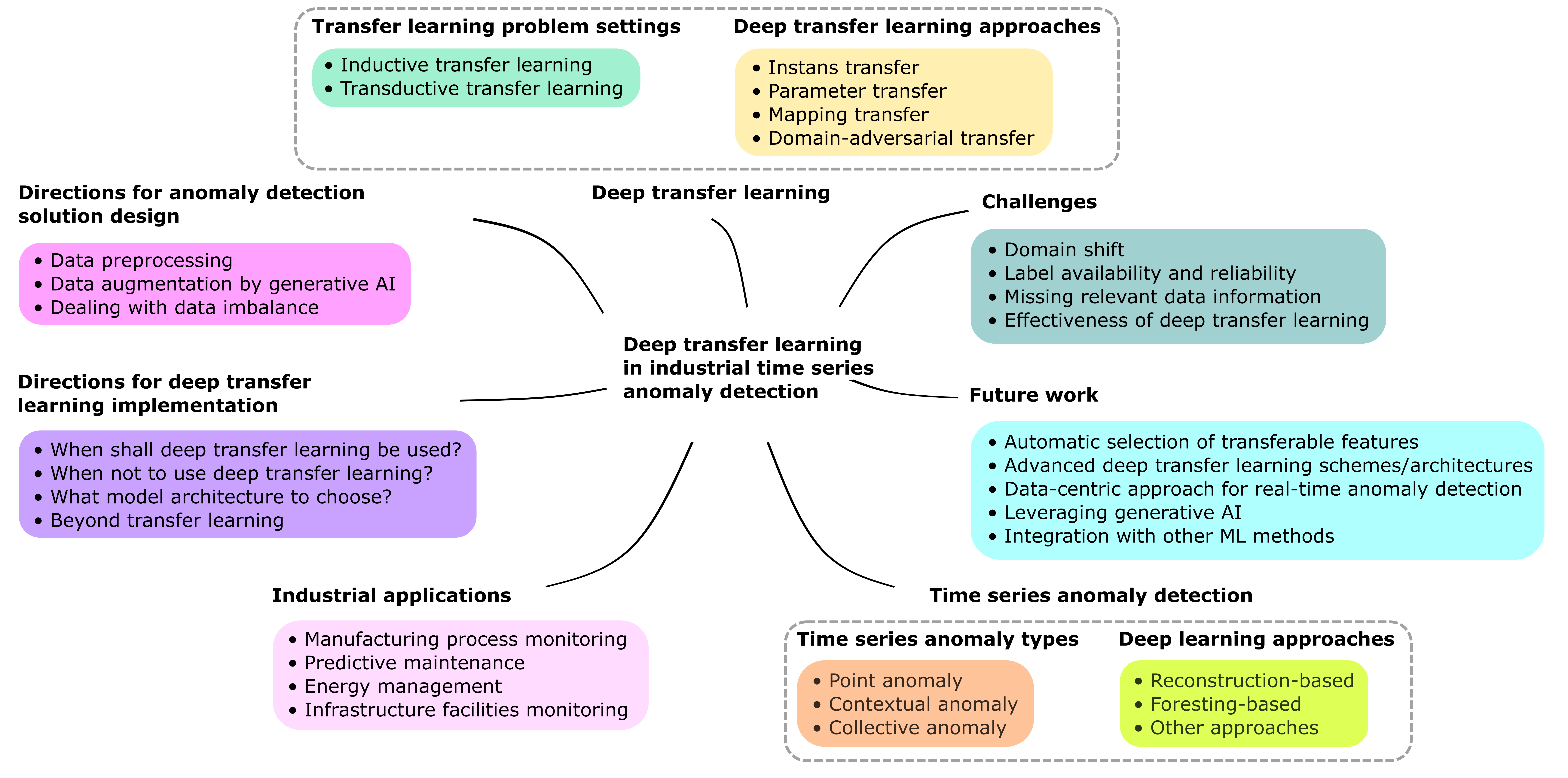}
\caption{\label{survey_taxomony}A generic taxonomy in this paper to analyze deep transfer learning for industrial time series anomaly detection.}
\end{figure*}

Fig. \ref{survey_taxomony} illustrates the taxonomy in this survey to categorize reviewed studies based on different aspects, including deep transfer learning, time series anomaly detection, industrial applications, current challenges, and future directions.


\section{Deep transfer learning}
\label{sec_tl}
\subsection{Overview of the field}
\begin{figure*}[th]
\includegraphics[width=1\textwidth]{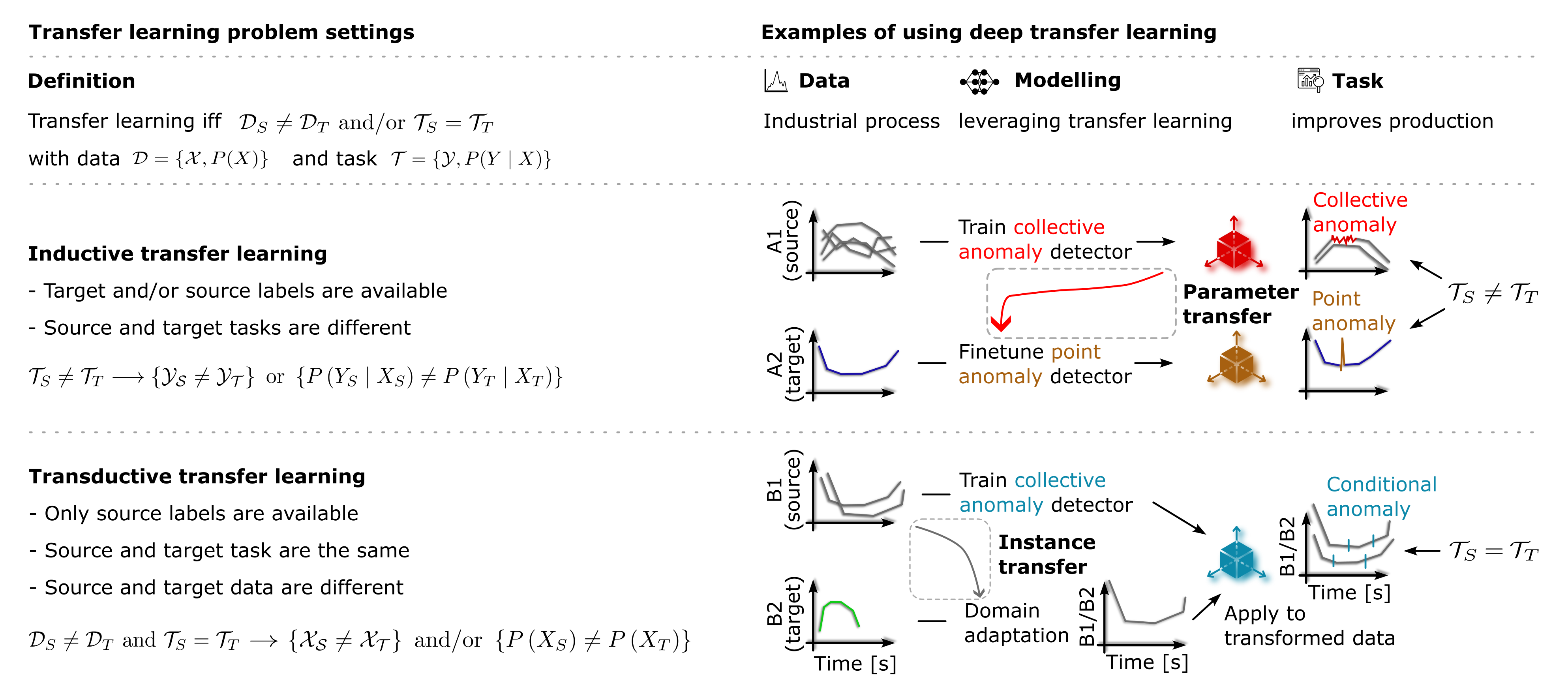}
\caption{Taxonomy of transfer learning problem settings (left; see Section \ref{sec:def} for the definition of terms) and corresponding examples using deep transfer learning approaches (right). 
On the left, we classify transfer learning problems as inductive or transductive transfer settings
Correspondingly, we provide two examples using deep transfer learning methods: 
In the inductive transfer setting, we collect time series data from screw production and wrench production. 
Labeled screw data (A1) is used to detect collective anomalies (a set of data points behaving differently compared to the entire time series \cite{choi2021deep, chandola2009anomaly}, further explained in Section \ref{sec_tsanomalydetection}).
Then, parameter transfer (Section \ref{sec_parameter_tl}) is applied to transfer knowledge by fine-tuning the pre-trained model from labeled screw data to detect point anomalies (further explained in Section \ref{sec_tsanomalydetection}) on labeled wrench data (A2).
For the transductive transfer setting in the lower panel, we present a different situation for contextual anomaly detection (further explained in Section \ref{sec_tsanomalydetection}). In this case, we have two datasets, B1 and B2, analyzed using the same model. However, the data in B2 significantly differs in appearance from the data in B1. To address this problem, instance transfer (further explained in Section \ref{sec_instance_tl}) is used.
Through this learning process, the data in B2 is transformed in a way that makes it compatible with the model that has been trained exclusively on data from B1. Transfer learning, in this case, is thus achieved by adapting the data to fit the model through domain adaptation rather than adjusting the model to fit new data.}
\label{taxonomy}
\end{figure*}

\change[]{Transfer learning in a deep learning setting aims to increase the efficiency, performance, and generalization of deep learning models by transferring
knowledge from one dataset and task to a new one.}
{Transfer learning in the setting of industrial time series analysis for anomaly detection is a tool to increase the flexibility of autonomous process monitoring.
It addresses the challenge of adapting the algorithm, and thus the decision process, to a related but previously unseen setting where limited training data is available.}
\change{This}{The transfer} eliminates the need to train a deep learning model from scratch, which in turn reduces the amount of necessary data and compute 
required to solve a new task or \add{adjust to a }new data domain.
In either case, knowledge is transferred from a source to a target domain, as \change{defined}{described} below.
The transfer learning problem settings can be categorized as inductive or transductive transfer depending on the data and task conditions. \change{, while we}{We} categorize deep learning-based transfer learning approaches into instance transfer, parameter transfer, mapping transfer and domain-adversarial transfer. We illustrate them by using two intuitive examples in Fig. \ref{taxonomy}, with more details being elaborated in the following sections.




\subsection{Formal description of deep transfer learning}
\label{sec:def}

Domain $\mathcal{D}$ includes the domain feature space $\mathcal{X}$ and marginal data distribution $P(X)$ as $\mathcal{D} = \{\mathcal{X}, P(X)\}$, where $X$ is the domain data, $X=\left\{x_{1}, \ldots, x_{n}\right\} \in \mathcal{X}$.
Similarly, a learning task is defined as $\mathcal{T} = \{\mathcal{Y}, f_{\mathcal{T}}(\cdot)\}$, where $\mathcal{Y}$ denotes the task space and usually represents class label. For anomaly detection tasks, $\mathcal{Y}$ is the set of the two classes ``normal'' and ``abnormal''. 
The function $f_{\mathcal{T}}(\cdot)$ can be used to predict the corresponding label of a new instance $x_i$.
The objective predictive function $f_{\mathcal{T}}(\cdot)$ learned from domain data can be interpreted as a form of conditional probability. Thus, the learning task can be rewritten as $\mathcal{T} = \{\mathcal{Y}, P(Y|X)\}$, where $P(Y|X)$ is used as a likelihood measure to determine how well a given data set $X$ fits with a corresponding class label set $Y$.

\label{sec_tlps}

In the surveyed literature on transfer learning for anomaly detection in industrial applications,
transfer learning methods from other fields, such as computer vision and natural language processing, were adopted.
We therefore use a generic classification scheme for transfer learning methods.
We largely follow the definition of transfer learning in literature \cite{zhuang2020comprehensive, pan2010survey}. Given a source domain $\mathcal{D}_{S}$ and learning task $\mathcal{T}_{S}$, as well as a target domain $\mathcal{D}_{T}$ and learning task $\mathcal{T}_{T}$, transfer learning aims to improve the performance of the predictive function $f_{\mathcal{T}}(\cdot)$ in $\mathcal{D}_{T}$  by transferring knowledge from $\mathcal{D}_{S}$ and $\mathcal{T}_{S}$, where $\mathcal{D}_{S} \ne \mathcal{D}_{T}$ and/or $\mathcal{T}_{S} \ne \mathcal{T}_{T}$. Usually, the size of source dataset is much smaller than target dataset.

This definition of transfer learning can be broadened, i.e., the target task can benefit from multiple source domains. 
Transfer learning is thus the idea of making the best use of related source domains to solve new tasks. 
In contrast, traditional machine learning (ML) methods learn each task separately from scratch, and each respective model can only be applied to the corresponding task.




We define a taxonomy of transfer learning problem settings as shown in Fig. \ref{taxonomy} mainly depending on the label availability in the two domains to be easily applicable to the requirements of a case at hand (compare different definitions for other purposes in the literature \cite{zhuang2020comprehensive} --  \cite{tan2018survey}).

We differentiate it into inductive and transductive transfer learning \cite{pan2010survey}.
Inductive transfer learning is applied when the target task is different from the source task, i.e., $\mathcal{T}_{S} \ne \mathcal{T}_{T}$ (meaning that $\{\mathcal{Y_{S}} \ne \mathcal{Y_{T}} \}$ or $\{P(Y_{S}|X_{S}) \ne P(Y_{T}|X_{T}) \}$). 
The conditional probability distribution is induced with labeled training data in the target domain \cite{torrey2009transferlearning}.
A corresponding example is illustrated as Scenario A in Fig. \ref{taxonomy}, where the learning tasks are different and the goal of transfer learning is to recognize point anomaly from the collective anomaly task.
Related areas of inductive transfer learning are multi-task learning \cite{caruana1997multitask, ruder2017overview} and sequential learning, depending on whether tasks are learned simultaneously or sequentially. 

Transductive transfer learning is applied when the source and target tasks are the same, while the source and target domain are different, i.e., $\mathcal{T}_{S} = \mathcal{T}_{T}$ and $\mathcal{D}_{S} \ne \mathcal{D}_{T}$ (meaning that $\{\mathcal{X_{S}} \ne \mathcal{X_{T}} \}$ or $\{P(X_{S}) \ne P(X_{T})\}$. A subcategory is domain adaptation \cite{Kouw2019} when the feature space of source and target data are the same but the corresponding marginal distributions are different (i.e., $\{\mathcal{X_{S} = X_{T}} \}$ and $\{P(X_{S}) \ne P(X_{T})\}$).
Scenario B in Fig. \ref{taxonomy} is an example of transductive transfer learning
where the learning tasks are identical, and the goal of transfer learning is to recognize
contextual anomalies in an unlabelled data set.



\subsection{Deep transfer learning approaches}
Since deep neural networks (DNNs) can learn useful feature representations from large amounts of data through back-propagation \cite{Bengio2014_Represen}, they have been widely adopted for tackling complex problems in practice \cite{schmidhuber2015deep, 7738816, inbook111, schmidhuber2022annotated}, which involve large-scale and high-dimensional data.
Deep transfer learning methods implement transfer learning principles within DNN
and, among other things, enable deep learning based analysis pipelines to be applied to new datasets.


Based on the transferring techniques in the surveyed literature, we access how knowledge is shared across domains and help increase the performance in the target task or domain. 
we divide deep transfer learning approaches further into $4$ categories: instance transfer, parameter transfer, mapping transfer, and domain-adversarial transfer, as illustrated in Table \ref{tab:dtl}. Furthermore, instance transfer, mapping transfer, and domain-adversarial transfer can be described as data-driven approaches. They focus on transferring knowledge by leveraging a large amount of data. It usually involves transforming and adjusting the data instances or manipulating data from different domains by feature alignment, feature mapping, etc. 
On the other hand, parameter transfer is a model-driven approach, which places more emphasis on understanding the underlying structure and dynamics of the data. It usually involves transferring the parameters of pre-trained model from source domain to target domain.

\def\arraystretch{1}\tabcolsep=12pt
\begin{table*}[t]

\caption{Overview of deep transfer learning approaches with references.}
\label{tab:dtl}

\centering

\fontsize{8}{12}\selectfont 
\begin{tabular}{p{4cm}p{7cm}p{2cm}}
\specialrule{1pt}{0pt}{0pt} 
\textbf{Deep transfer learning approach}     &  \textbf{Short description}                  & \textbf{References}               \\
\hline\hline
Instance transfer     & Augmenting target data by transforming data instance from the source domain to the target domain               & \cite{he2022instance, amirian2021prepnet, wang2019instance}              \\
\hline
Parameter transfer     & Transfering learned parameters of a pre-trained model from source domain and adapting the model for target domain                   & \cite{bommasani2021opportunities, Bert2019, zhangAnomalyDetectionControl2023, vaswani2017attention, dou2021gpt, alexandr2021fine, weberDetectionBuildingOccupancy2020, Guo2019SpotTune,sager2022unsupervised, NEURIPS2020_1457c0d6}               \\[2mm]
\hline
Mapping transfer     & Reducing feature discrepancies between source and target domains by minimizing the distance between mapped features in the latent space  & \cite{zhuang2020comprehensive, wangFewShotTransferLearning2022, tzeng2014deep, long2017deep, long2015learning, zhang2015deep, venkateswara2017deep}\\
\hline
Domain-adversarial transfer    & Extracting an indiscriminative feature representation between source and target domain through adversarial training                    & \cite{soleimaniCrosssubjectTransferLearning2021,tzeng2015simultaneous, ozyurtContrastiveLearningUnsupervised2023, ganin2016domain, ajakan2014domain, Tzeng_2017_CVPR}\\
\specialrule{1pt}{0pt}{0pt} 
\end{tabular}

\end{table*}





\subsubsection{Instance transfer}
\label{sec_instance_tl}
The intuition of instance transfer is that although source and target domains differ, it is still possible to transform and reuse source data together with a few labeled target samples. 
A typical approach is to re-create some labeled data from the source domain.
For example, He \textit{et al.} propose an instance-based deep transfer learning model with an attention mechanism to predict stock movement \cite{he2022instance}. They first create new samples from the source dataset that are similar to the target samples by using an attention network and then train another network on the created samples and target training samples for prediction tasks. Since two networks are trained separately for different tasks, it needs further investigation to what extent the generated samples can contribute to the prediction task. 
Amirain \textit{et al.} introduce an innovative instance transfer method for domain adaptation \cite{amirian2021prepnet}. 
They propose an effective auto-encoder model with a pseudo-label classifier to reconstruct new data instances that obtain general features across different datasets for medical image analysis. 
Taking another avenue, Wang \textit{et al.} exclude the source data that negatively impacts training target data. Specifically, they choose a pre-trained model from a source domain, estimate the impact of all training samples in the target domain, and remove samples that lower the model's performance. Then, the optimized training data is used for fine-tuning. The experiments are conducted on large image datasets \cite{wang2019instance}. Instead of transferring the data, the approach excludes certain samples based on the pre-trained model's predictions. Additional validation is required in industrial environments, especially when only a few data are available in some industrial settings. 

\subsubsection{Parameter transfer} 
\label{sec_parameter_tl}
Parameter transfer adapts the learned parameters of a pre-trained model to a new model. 
This assumes that DNNs can get similar feature representations from similar domains. 
Thus, through transferring parts of the DNN layers together with pre-trained parameters and/or hyperparameters, the pre-trained model is used as a base model to further train on target domain data and solve different learning tasks. 
Particularly, parameter transfer has gained popularity in computer vision and natural language processing, where large models are pre-trained on large datasets \cite{bommasani2021opportunities}.
In natural language processing, for example, BERT \cite{Bert2019} and GPT-3 \cite{NEURIPS2020_1457c0d6} are based on the Transformer architecture \cite{vaswani2017attention} which can be fine-tuned for a variety of downstream tasks, including content generation \cite{dou2021gpt}, language translation \cite{sun2021multilingual}, question answering \cite{glass2019span}, and summarization \cite{alexandr2021fine}. 
In computer vision, Yosinski \textit{et al.} investigate the general transferability of CNNs in image recognition \cite{yosinski2014transferable}. They analyze the transferring effect by fine-tuning or freezing a certain amount of layers in the networks. Experimental results show that transferring features from source to target domain improves network generalization compared to those trained solely on the target dataset. Additionally, they quantify the model performance by assessing how features at what layers transfer from one task to another. It is surprising to find that transferring a pre-trained network from any number of layers can produce a boost for fine-tuning on a new dataset. However, the experiments are only conducted on certain image datasets, and Tuggener \emph{et al.} show \cite{tuggener2021imagenet} the limits of parameter transfer when the chosen architecture is overfitted on the particularities of certain large-scale datasets.

Unlike the typical way of fine-tuning a pre-trained model, Guo \textit{et al.} propose an adaptive fine-tuning approach SpotTune to find the optimal fine-tuning strategy for the target task \cite{Guo2019SpotTune}. Specifically, a policy network is used to make routing decisions on whether to pass the target instance through the pre-trained model. The results show SpotTune is effective in most cases by using a hybrid of parameter and instance transfer.
Sager \textit{et al.} propose an unsupervised domain adaptation for vertebrae detection in 3D CT volumes by transferring knowledge across domains during the training process \cite{sager2022unsupervised}. 


\subsubsection{Mapping transfer}
\label{sec_map_tl}
Mapping transfer refers to learning a related feature representation for the target domain by feature transformation, which includes feature alignment, feature mapping, and feature encoding \cite{zhuang2020comprehensive}. The goal is to reduce feature discrepancies between source and target domains by minimizing the distance between the distribution of latent feature representation.
There are various criteria to measure the distribution difference, including Wasserstein distance \cite{shenWassersteinDistanceGuided2018}, Kullback-Leibler Divergence \cite{daiCoclusteringBasedClassification2007}, etc. Among them, Maximum Mean Discrepancy (MMD) \cite{tzeng2014deep} is most frequently adopted in mapping transfer from the surveyed papers. 
The MMD is calculated as the difference between the mean embeddings of the samples in a reproducing kernel Hilbert space associated with a chosen kernel function.
Added to the target loss function, it serves as a powerful tool for comparing the similarity of complex, high-dimensional datasets using a wide variety of kernel functions. 

Previous work has focused on transferred feature extraction/dimensionality reduction using MMD. 
Wang \textit{et al.} focus more on the subdomain of the same subcategory instead of the alignment of the global distribution between source and target domain \cite{wangFewShotTransferLearning2022}. Specifically, they first use the attention mechanism to extract discriminative features that are most related to the fault signal. Then, local MMD is applied to transfer knowledge to adjust the distribution of related subdomains under the same category. 
Long \textit{et al.} propose their Joint Adaptation Network \cite{long2017deep} based on MMD, in which the joint distributions of multiple domain-specific layers across domains are aligned. In addition, an adversarial training version was adopted to make distributions of the source and target domains more distinguishable.   
Similarly, Long \textit{et al.} adopt multi-layer adaptation and proposed Deep Adaptation Networks (DAN) \cite{long2015learning}. The first three convolutional layers are used in DAN models to extract general features. For the last three layers, multi-kernel MMD bridges the cross-domain discrepancy and learns transferable features. 
Zhang \textit{et al.} propose a Deep Transfer Network in which two types of layers are used to obtain domain invariant features across domains by adding MMD loss. The shared feature extraction layers learn a shared feature subspace
between the source and the target samples, and the discrimination layer is then used to match conditional distributions by classifier transduction \cite{zhang2015deep}.
Venkateswara \textit{et al.} propose Deep Adaptation Hash network \cite{venkateswara2017deep}, which is fine-tuned from the VGG-F \cite{chatfieldReturnDevilDetails2014} network. Multi-kernel MMD loss is employed to train
the Deep Adaptation Hash network to learn feature representations that align the source and target domains. 

\subsubsection{Domain-adversarial transfer}
Inspired by Generative Adversarial Networks (GANs) \cite{goodfellow2020generative, SCHMIDHUBER202058}, the goal of domain-adversarial transfer is to extract a transferable feature representation that is indiscriminative between source and target domain through adversarial training.
Adversarial transfer is primarily concerned with addressing domain adaptation problems. 

Soleimani and Nazerfard utilize the GANs framework to perform cross-subject transfer learning \cite{soleimaniCrosssubjectTransferLearning2021}. The generator is used to generate samples that are similar to the target data. Meanwhile, the discriminator distinguishes the fake samples from the target samples.
The classifier is trained to discriminate the labeled source data and fake samples to learn generalized features invariant to source and target domains. It is important to note that in real-world applications, training GANs can be unstable due to mode collapse, especially in the case when the source data and target data are unbalanced, and the generator may fail to generate fake samples that can confuse the discriminator. 
Tzeng \textit{et al.} adopt a domain confusion loss across the source and target domains to learn a domain invariant representation \cite{tzeng2015simultaneous}.
Ganin \textit{et al.} propose a new domain adaptation architecture by adding a domain classifier after feature extraction layers \cite{ganin2016domain}. A gradient reversal layer is used to ensure the similarity of the feature distributions over source and target domains. 
Similarly, Ozyurt \textit{et al.} develop a novel framework for unsupervised domain adaptation of time series data by using contrastive learning and domain-adversarial transfer learning \cite{ozyurtContrastiveLearningUnsupervised2023}. A domain classification loss is applied to extract domain invariant features. The drawback is that the experiments are designed in a way that the source and target data sizes are similar, whereas in practice, the target data is usually much fewer than the source data.
Ajakan \textit{et al.} propose a domain adversarial DNN in which a domain regressor is applied to learn a domain invariant feature representation \cite{ajakan2014domain}. 
Tzeng \textit{et al.} use an unsupervised domain adaptation method that combines adversarial learning with discriminative feature learning \cite{Tzeng_2017_CVPR}.\\[4mm]

\subsection{Related learning paradigms}
Besides the dedicated transfer learning approaches discussed above, there are methods that represent alternative ways to solve tasks across domains or are complementary to the native transfer learning methods.

\begin{itemize}
\item \noindent\textit{Multi-task learning} is a machine learning technique where a single model is trained on multiple tasks simultaneously. The idea is to improve the performance of the model by learning a shared representation that captures the features between all tasks.
Because the network learns to solve multiple tasks, it may generalize better to new data and tasks.

\item \noindent\textit{Continuous learning} \cite{parisi2019continual} is a learning process where the model continuously learns new tasks from previous tasks over time without forgetting how to solve previous tasks. To some extent, continuous learning can be seen as a sequential transfer learning process, with the constraint to preserve the performance of the previous tasks, which leads to an accumulation of knowledge over time.

\item \noindent\textit{Few-shot learning} \cite{wang2020generalizing} is a type of machine learning where a model can learn and perform well on a new task with only a limited number of labeled samples. In extreme cases, the model can learn with one label \cite{fei2006one} and without any label \cite{lampert2009learning}. 
Whereas, transfer learning usually involves reusing the model from relevant tasks and continuing training on the target dataset.

\item \noindent\textit{Domain generalization}
\cite{zhouDomainGeneralizationSurvey2023, blanchardGeneralizingSeveralRelated2011} focuses on developing a generalized model from one or multiple distinct domains to detect unseen target domain data. The main goal is to overcome the domain shift problem. Domain generalization and transfer learning are both applied to transfer knowledge from source domain to target domain. The major difference between transfer learning and domain generalization lies in the utilization of target domain data. Transfer learning leverages knowledge from source domain and target domain. In contrast, domain generalization solely learns from source domain, without access to the target data. 

\item \noindent\textit{Meta-learning} \cite{finn2017model, hospedales2021meta} is known as “learning to learn”. For meta-learning, models are trained on a different set of tasks instead of a set of data in the traditional machine learning setting. In this sense, meta-learning can be seen as a form of transfer learning because it involves transferring knowledge from task to task.

\item \noindent\textit{Knowledge distillation} \cite{gouKnowledgeDistillationSurvey2021} effectively learns a small model trained to mimic the behavior of a larger, more complex
model. The knowledge learned by the larger model can be transferred to the smaller model, which can then be used for the target task.

\item \noindent\textit{Self-supervised learning} \cite{liuSelfSupervisedLearningGenerative2023, bai2021self} involves training a model to predict some aspect of the input data without
any external supervision. The learned representations can be used for various downstream tasks, including those
that involve transferring knowledge from one domain to another.
\end{itemize}

\section{Time series anomaly detection in industry}
\label{sec_tsanomalydetection}

Time series anomaly detection encompasses statistical techniques to analyze and interpret sequential temporal data.
In the context of industrial processes, time series anomaly detection plays a crucial role in automating monitoring, effectively scheduling maintenance, and controlling the efficiency, quality, and performance of these processes.
For example, after the detection of an anomaly, another model that captures the relationship between
time course and different failure modes or drifts may be exploited for predictive maintenance.
For example, in injection molding process monitoring, anomaly detection models are used to analyze recorded sensor data from injection molding machines to detect bad parts and identify the root cause of anomalies \cite{Tercan2018bridging}.
There are two basic ways to detect anomalies: for supervised anomaly detection, labels (normal/abnormal) are needed per time series to build a binary classifier \cite{gornitz2013toward}.
For unsupervised anomaly detection, an anomaly score or confidence value that is conditioned purely on normal data can be used to differentiate abnormal from normal instances \cite{zhang2021unsupervised} --\cite{stadelmann2019beyond}.



\subsection{Anomaly types}
\begin{figure}[t!]
  \centering
    \includegraphics[width=0.6\columnwidth]{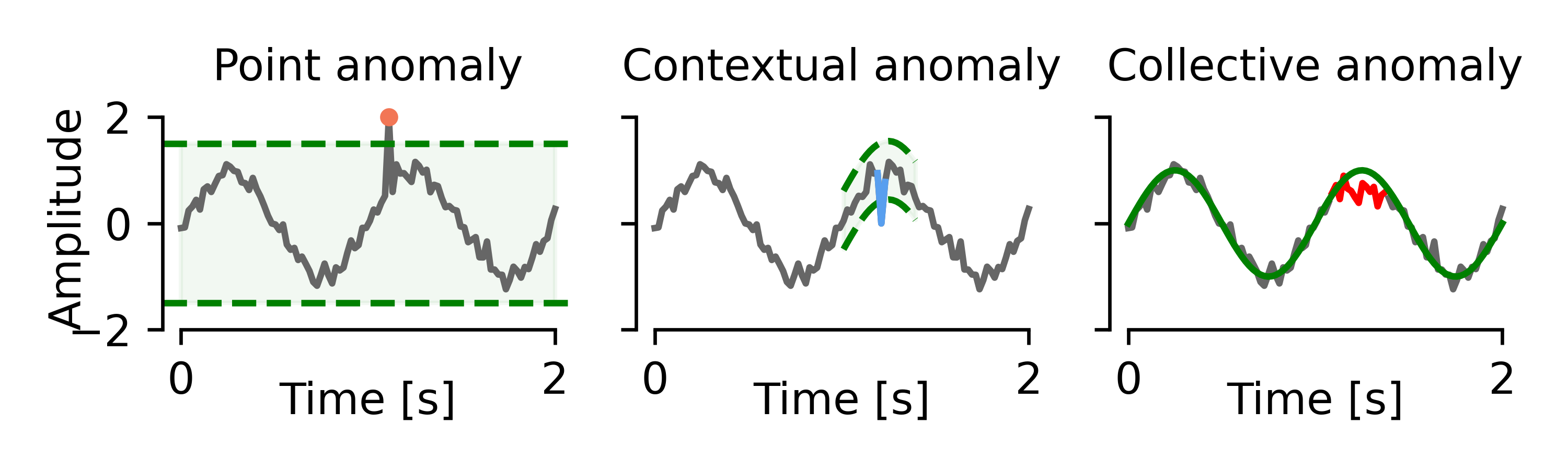}
   \caption{Three time series anomaly types. Gray lines represent recorded time series signals, and dashed green lines are \textsl{a priori} set thresholds of normal operations. The red dots and the red line indicate anomalies. Point anomalies are single values that fall outside of a pre-set range (left panel). Contextual anomalies are samples that deviate from the current context (middle panel). Collective anomalies are defined as a series of data points that all fall within the range of operation but jointly are not expected (right panel).}
  \label{tsad}
\end{figure}

According to the literature \cite{hawkins1980identification}, an outlier is an observation that deviates significantly from other observations in a way that it is likely that it was generated by a different mechanism.
In this survey, we focus on time series data collected from machine sensor readings in the context of industrial applications, either univariate (only one variable is recorded over time) or multivariate (several simultaneously recorded measurements). 
Time series anomalies might occur for various reasons, including internal factors (e.g., temporary sensor error, machinery malfunction) and external factors (e.g. human error, ambient temperature). They can be divided into three categories \cite{choi2021deep}, \cite{chandola2009anomaly}: point anomalies, contextual anomalies, and collective anomalies.
Point anomalies are isolated samples that deviate significantly from the normal behavior of that time series, which can be seen on the left of Fig.~\ref{tsad}, e.g., a sudden spike in a pressure reading from a manufacturing machine sensor. These point anomalies can be caused by temporal sensor error, human error, or abnormal machinery operations. 
Contextual anomalies represent data points that deviate from normal ones only in their current context, and an example can be seen in the middle of Fig.~\ref{tsad}.  
Collective anomalies are a set of data points that in their entirety (but not individually) are abnormal with respect to the entire time series, as shown on the right of Fig.~\ref{tsad}. 

\subsection{Challenges}
Challenges regarding detecting time series anomalies persist due to two specific properties:
(1) The \textit{complexity of time series data}.
As the automation level of industrial processes and the complexity of industrial systems increases, univariate time series data become insufficient and inefficient in representing any industrial process in its entirety. 
Hence, more sensors are installed to monitor the whole process, making it necessary to detect anomalies from multivariate time series, which poses particular challenges since it requires consideration of temporal dependencies and relationships between variables and modalities. Many researchers work on discovering generalized patterns from spatial and temporal correlated multivariate time series data. Zhang \textit{et al.} propose a Deep Convolutional Autoencoding Memory network \cite{zhang2021unsupervised}, where they build an autoencoder to capture spatial dependency of multi-variant data using MMD to distinguish noisy, normal and abnormal data. 
Zhu \textit{et al.} propose an interpretable model agnostic multivariate time-series anomaly detection method for applications of cyber physical systems \cite{zhuInterpretableMultivariateTimeseries2023}. The new method considers both the temporal and feature dimensions through an adaptive mask based series saliency module to produce accurate anomaly detection results and reasonable interpretations in the form of a mask matrix.
(2) The \textit{dynamic variability in industrial processes}. Industrial processes often have high dynamic variability and can be affected by a wide range of conditions, such as changes in temperature, pressure, and humidity. These conditions can cause fluctuations in the process outputs, which leads to data shift and domain shift. This can make it challenging to detect anomalies and maintain control over the industrial process.

\subsection{Anomaly detection methods}
\begin{figure}[t]
\centering
\includegraphics[width=0.6\columnwidth]{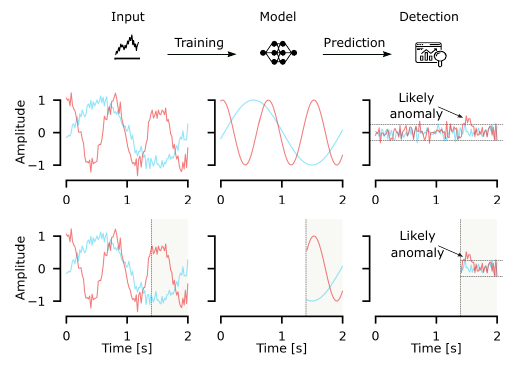}
\caption{\label{tsad_methods}
Illustration of deep learning-based anomaly detection (top row) with reconstruction-based (center row) and forecasting-based (bottom row) anomaly detection in time series.
The first column represents two time series.
The second column shows the reconstructed (top) and forecasted (bottom) time series.
The third column shows the difference between reconstructed/forecasted time series.
Deviations from the reconstructed or forecasted time series are indicative of an anomaly.
In deep reconstruction-based anomaly detection, the entire sequence is reconstructed in a decoder-encoder architecture, and the reconstructed sequence is used to compare with the actual sequence. In deep forecasting-based anomaly detection, the end of a sequence is predicted
using the start of the sequence and only the forecasted sequence is used to assess
the similarity to the ground truth.
In this example, the red time series has a likely anomaly at about 1.5 seconds.
(Best viewed in color.)
}

\end{figure}

Time series anomaly detection has been investigated for decades, and various types of methods have been proposed \cite{schmidlAnomalyDetectionTime2022a}. This paper exclusively discusses the time series anomaly detection techniques using deep learning, leveraging its robust representation learning capabilities. Current deep learning methods can be mainly divided into reconstruction-based, forecasting-based, and other methods. Fig.~\ref{tsad_methods} illustrates the two main methods. In deep reconstruction-based anomaly detection, the reconstructed sequence is used to compare with the actual sequence. Differently, in deep forecasting-based anomaly detection, only the forecasted sequence is used to assess the similarity to the ground truth.


\paragraph{Reconstruction-based methods}
Reconstruction-based methods aim to learn the data distribution of the normal time series and differentiate the abnormalities from the normal ones by computing the reconstruction errors. Audibert \textit{et al.} propose a fast and stable method -- Unsupervised anomaly detection for multivariate time series \cite{audibert2020usad}, based on adversely trained autoencoders. The encoder-decoder architecture within an adversarial training framework combines the advantages of autoencoders and adversarial training while compensating for the limitations of each technique. After training two autoencoders, the anomaly score is defined by balancing the reconstructed errors from the two autoencoders with two hyperparameters. However, the challenge arises in selecting these two hyperparameters of the anomaly score when the testing dataset is unavailable. 

Malhotra \textit{et al.} also formulate an anomaly score based on reconstruction error \cite{malhotraLSTMbasedEncoderDecoderMultisensor2016b}. They first train the LSTM encoder-decoder model to reconstruct the normal time series. Subsequently, they leverage the reconstruction errors to calculate the probability by using Maximum Likelihood Estimation to detect a specific point within a time series as an anomaly. They set a window to detect anomalies. The window will be labeled as anomalous if the probability exceeds a threshold. 
Similarly, Wei \textit{et al.} also propose an LSTM-based encoder-decoder model to detect multivariant time series sequences based on the reconstruction error \cite{weiLSTMAutoencoderBasedAnomalyDetection2023}. The major difference is the anomaly detection criteria. They assume the reconstruction error of train/test data follows the normal distribution and detect anomalies by using the 2-sigma rule of the normal distribution as a threshold.
Zeng \textit{et al.} propose an adversarial transformer structure to detect multivariate time series anomalies effectively \cite{zengMultivariateTimeSeries2023}. Here, two-stage adversarial training is applied for the transformer. In the first stage, two transformers are trained by minimizing the reconstruction error to capture the temporal trends in the time series. In the second stage, the reconstruction error serves as prior knowledge in the adversarial training process, enabling the model to distinguish anomalies from normal time series. Then, an anomaly score is defined by combining the anomaly probability and reconstruction error. Again, a threshold has been chosen to differentiate anomalies from normal ones.

GANs, as effective unsupervised learning methods, have been used in time series anomaly detection. Anomaly detection methods based on GANs focus on extracting features by adversarial training on normal samples. Consequently, features from the abnormal samples diverge from those of the normal ones, reflecting in reconstruction error and discrimination value.
Li \textit{et al.} use LSTM-RNN as a base model for building generator and discriminator in GAN\cite{liMADGANMultivariateAnomaly2019}. 
The proposed framework considers multiple variables to capture the temporal correlation of multi-time series distributions. Additionally, they proposed a novel anomaly score, which can detect anomalies through discrimination and reconstruction. 
More specifically, the score is a combination of the reconstruction difference between generated data and original data and the discrimination results from the discriminator.  
Similarly, Niu \textit{et al.} and Bashar \textit{et al.} both propose an LSTM-based VAE-GAN for time series anomaly detection, where LSTM networks are used as the generator, and discriminator \cite{niuVAEGANTimeSeries2020, basharTAnoGANTimeSeries2020}. When it comes to anomaly scores, setting an optimal threshold is usually a critical step. However, using a small portion of the test set to decide the optimal threshold may not be practical in real-world scenarios \cite{niuVAEGANTimeSeries2020}. Additionally, it is important to note that the method has been only tested for point anomaly detection, further investigation is required when they are applied to detect other anomaly types.

\paragraph{Forecasting-based methods}
Forecasting-based methods predict the value of the following timestamps and predict temporal anomalies according to the prediction error.
Kim \textit{et al.} propose a forecasting-based unsupervised time-series anomaly detection method using transformer architecture \cite{kimTimeseriesAnomalyDetection2023}. The idea is to train a transformer-like model by forecasting a fixed-length time series based on the previous timestamps. The trained model is used to predict time series with an anomaly score such that an instance where the anomaly score is larger than a static threshold is defined as an anomaly. A dynamic thresholding technique is also mentioned but not explicitly discussed in the paper. 

Deng and Hooi propose a novel attention-based graph neural network approach \cite{dengGraphNeuralNetworkBased2021} that learns a graph of dependence relationships between multi-variant time series signals by forecasting the behavior based on past time series. Then, a graph deviation scoring is defined for each sensor to detect and explain anomalies. 
Tang \textit{et al.} propose an interpretable multivariate time series anomaly detection method based on graph neural networks and gated recurrent units \cite{tangGRUBasedInterpretableMultivariate2023}. The feature representation is learned through forecasting the future time series segment. An abnormal score is set for each time series to detect anomalies. The feature embedding is then used for 2D visualization through t-SNE plots to interpret the clusters within time series from different sensors.

\paragraph{Other methods}
Ding \textit{et al.} propose a joint network to integrate the advantages of reconstruction and forecasting/prediction \cite{dingMSTGATMultimodalSpatial2023}. First, they propose a multimodal graph attention network to tackle the spatial-temporal dependencies for multimodal time series. Further, they optimize the reconstruction and prediction modules simultaneously to predict anomalies. 
Himeur \textit{et al.} take advantage of annotated data and directly use a DNN as a classifier to classify normal and abnormal energy consumption types  \cite{himeurNovelApproachDetecting2020}. The enormous imbalance of real anomaly patterns is one concern in the approach. Thus, a normalized technique of power consumption data is applied to deal with this problem. The normalized data represent the difference in power consumption rates of each current time sample and the previous one. It can provide information on how fast the consumption reacts to the time evolution.
However, any further evaluation of this technique is not discussed, and it is still an open problem regarding anomaly detection for other datasets. 
Yang \textit{et al.} propose a contrastive learning structure with dual attention to learn a permutation invariant representation of the data with superior discrimination characteristics between normal points and anomalies \cite{yangDCdetectorDualAttention2023}. Unlike most reconstruction-based models, their model is a self-supervised framework based on representation learning. The new method achieves state-of-the-art comparable performance on six multivariate and one univariate time series anomaly detection benchmark datasets. However, the extensive framework with two multi-head-attention blocks may be prone to overfitting. This concern is amplified by the absence of training details, leaving only evaluation details disclosed.\\

In principle, these anomaly detection approaches are applicable to all types of anomalies. Reconstruction-based methods are typically applied to the entire or a portion of the time series. Long-time series are commonly segmented into subsequences using a predefined sliding window. In the case of detecting context/collective anomalies, the reconstructed loss of the time series sequence is evaluated within the predefined sliding window, if the reconstruction loss is larger than an acceptable threshold, then that time series sequence is classified as an anomaly. In the case of point anomalies, reconstruction is performed at each single time stamp, akin to a regression problem, and then the reconstruction loss of each single timestamp is evaluated to determine whether the single timestamp is anomalous or not. It is also applied to forecasting-based anomaly detection methods, instead of computing reconstruction error, forecasting-based methods predict the value in the next time stamp for point anomalies or the next time series sequence for context/collective anomalies. The anomalies will be detected based on the deviation between the predicted value and the normal value. Other anomaly detection approaches usually combine the reconstruction-based and foresting-based methods.  

To sum up, these methods are applicable to each type of anomaly. However, the effectiveness of these anomaly detection approaches may vary depending on the anomaly detection tasks at hand, which are characterized by the granularity at which the time series data is observed and analyzed. 

\section{Industrial applications}
\label{sec_industrialapplications}
\subsection{Overview}
Deep transfer learning techniques have gained prominence in computer vision and natural language processing, primarily due to the abundance of available datasets.
However, their adoption in the context of industrial time series data has been comparatively limited. This hesitancy can be attributed to the limited public availability of such datasets and the unique domain-specific characteristics they possess, which complicate generalized advancements.
Encouragingly, there has been a recent uptick in the application of deep transfer learning for anomaly detection within the industry such as fault diagnosis \cite{li2022perspective}, quality management \cite{ma2019improving}, manufacturing process monitoring \cite{Tercan2018bridging}, network/software security \cite{rosenberg2018end}, 
 and infrastructure monitoring \cite{pan2023transfer}. These can be mapped onto the core industrial domains of manufacturing process and infrastructure monitoring, predictive maintenance, and energy management. Table \ref{tab:industrial application} presents a compact comparison of
the related works using deep transfer learning approaches to solve these tasks. 

\begin{figure}[t]
 \centering
  \includegraphics[width=0.6\textwidth]{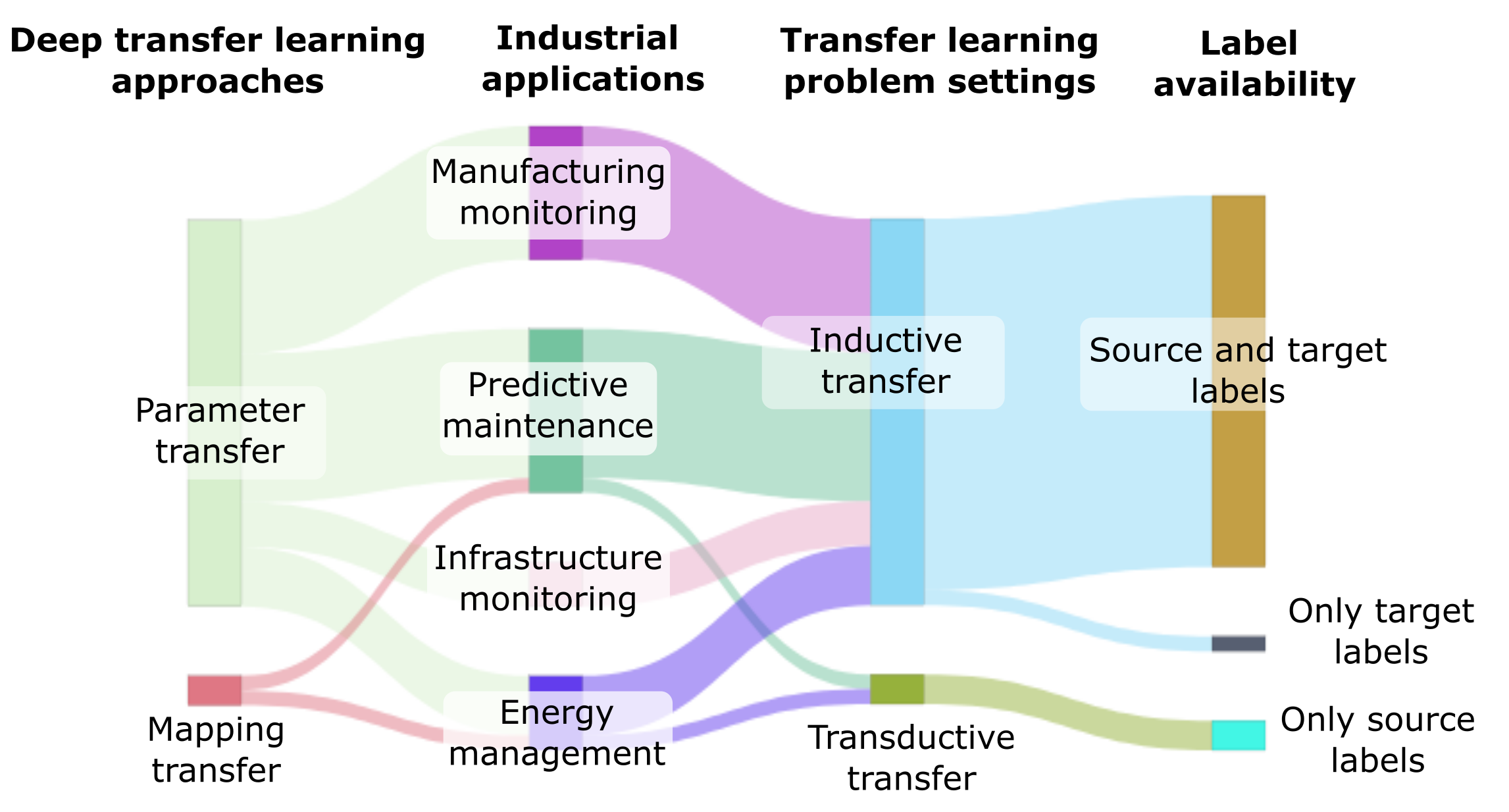}
  \caption{Overview of Sankey diagram of transfer learning problem setting, deep transfer learning approach categories, and label availability in the surveyed industrial domains.}
  \label{streamflow}
\end{figure}

Fig. \ref{streamflow} illustrates the Sankey diagram of the connections between industrial applications and the deep transfer learning approaches based on our literature survey. 
The diagram shows every path that connects the four dimensions of the methodology-problem-landscape within the surveyed literature. The broader the path is, the more papers are related to the linked topics. The goal is to give an overview of how deep transfer learning is applied to industrial problems in the recent literature and specifically show with these four dimensions: (1) which deep transfer learning approaches are actually used in practice; (2) what the main industrial domains for time series anomaly detection are; (3) what deep transfer learning category these domains belong to;  (4) what labels are available in source and target domain.


Key observations from Fig. \ref{streamflow} are:
(1) parameter transfer is much more frequently used than any other deep transfer learning approach across all surveyed industrial applications since fine-tuning a pre-trained model on target data is more straightforward to implement by taking advantage of the pre-trained model on the source dataset and usually without fundamental modification on the model architecture. It is noteworthy that instance transfer and adversarial transfer do not appear in the diagram. Apparently, these two deep transfer learning approaches are not considered effective in time series anomaly detection tasks in industry. 
The difficulty lies in implementing and training these scarcely researched approaches in the industrial field, as indicated by the findings.
(2) Hybrid approaches of parameter and mapping transfer can be seen in predictive maintenance. 
(3) Most industrial applications use inductive transfer learning, indicating they focus on leveraging labeled source and target data to solve the target task, i.e., use supervised learning.

\begin{sidewaystable*} 
\caption{A compact overview of industrial applications that used deep transfer learning for time series anomaly detection.}
\label{tab:industrial application}
\centering
\tiny

\begin{tabular}{p{0.5cm}p{4.2cm}p{2.5cm}p{1.5cm}p{1.5cm}p{1.5cm}p{1.5cm}p{0.5cm}p{0.5cm}}
\specialrule{0.6pt}{0pt}{0pt} 
Reference                         & Industrial task                    & Industrial domain                    & Deep transfer learning approach                                     & Transfer learning problem setting                   & Deep learning framework         & Source type & Source label                                          & Target label            \\
\hline\hline
\cite{maschler2021towards}           & Industrial metal forming anomaly detection                       & Predictive maintenance                                     & Parameter transfer                              & Inductive     & CNN                & Multiple & \checkmark                           &  \checkmark                 \\
\hline
\cite{wen2019time}                 & Monitoring systems Anomaly detection                & Predictive maintenance                                     & Parameter transfer                              & Inductive     & U-Net                  & Multiple &  \checkmark                            &  \checkmark                 \\
\hline
\cite{xu2019digital}                & Car body-side production line fault diagnosis                      & Predictive maintenance            & Parameter transfer                 & Inductive     & SAE                    & Multiple &  \checkmark       &  \checkmark            \\
\hline
\cite{mao2020robust}                 & Rotation bearings fault detection             & Predictive maintenance                                     & Mapping transfer                                                 & Transductive & Auto-encoder        & Multiple &  \checkmark                           & \xmark              \\
\hline 
\cite{wang2021anomaly}              & Industrial control systems anomaly detection                      & Predictive maintenance                                     & Parameter transfer                             & Inductive     & ResNet8                & Single  &  \checkmark                            & \checkmark                  \\
\hline
\cite{canizoMultiheadCNNRNN2019}      & Service elevator fault detection                         & Predictive maintenance                                     & Parameter transfer                             & Inductive     & CNN, RNN                & Multiple &  \checkmark                           & \checkmark                  \\
\hline
\cite{yao2022model}            & Nuclear power plants fault detection                   & Predictive maintenance                                     & Parameter transfer                             & Inductive     & CNN                    & Multiple &  \checkmark                            & \checkmark                 \\
\hline
\cite{ComparativeStudyDeep} & Building energy systems fault diagnosis & Predictive maintenance              & Parameter transfer  & Inductive     & CNN                    & Multiple &  \checkmark                             &  \checkmark                  \\
\hline
\cite{serradilla2021adaptable}    &  Press machine production prediction      & Predictive maintenance               & Parameter transfer                & Inductive     & CNN & Single  &  \checkmark             &  \checkmark                \\
\hline
\cite{zgraggen2021transfer}         & Wind turbine fault detection                       & Predictive maintenance                 & Parameter transfer                             & Inductive     & CNN                    & Single  &  \checkmark                               & \checkmark                \\
\hline
\cite{zabin2022hybrid}               & Industrial machine operating fault detection                 & Predictive maintenance                                     & Parameter transfer                             & Inductive     & CNN, LSTM               & Single  & \checkmark                      &  \checkmark                \\
\hline
\cite{liao2021manufacturing}         &  Machine turning operations classification             & Manufacturing process monitoring                           & Parameter transfer                             & Inductive     & VGG, ResNet              & Single   & \checkmark                                  &  \checkmark                  \\
\hline
\cite{lockner2021induced}          &  Injection molding process quality control                     & Manufacturing process monitoring                           & Parameter transfer                              & Inductive     & FCN                    & Multiple &  \checkmark           &  \checkmark                 \\
\hline
\cite{Tercan2018bridging}              & Injection molding process quality control                    & Manufacturing process monitoring                           & Parameter transfer                              & Inductive     & FCN                    & Single  & \checkmark            &  \checkmark                 \\
\hline
\cite{tercan2019industrial}          & Injection molding process anomaly detection     & Manufacturing process monitoring                           & Parameter transfer                             & Inductive     & FCN                    & Multiple & \checkmark         & \checkmark                 \\
\hline
\cite{lockner2022transfer}         & Injection molding process anomaly detection        & Manufacturing process monitoring                           & Parameter transfer                             & Inductive     & FCN                    & Multiple &  \checkmark          &  \checkmark                \\
\hline
\cite{gellrichDeepTransferLearning2021}  &  Aluminum gravity die casting quality prediction                                & Manufacturing process monitoring      & Parameter transfer                             & Inductive     & FCN                    & Single  &  \checkmark            &  \checkmark                  \\
\hline
\cite{abdallah2021anomaly}            &  Agriculture/manufacturing systems anomaly detection & Manufacturing process monitoring  & Parameter transfer                             & Inductive     & LSTM                   & Single  &  \checkmark                   &  \checkmark                  \\
\hline
\cite{abdallah2023anomaly}           &  Manufacturing testbeds anomaly detection & Manufacturing process monitoring  & Parameter transfer                    & Inductive     & LSTM, RNN & Single  &  \checkmark                              &  \checkmark                \\
\hline
\cite{maschler2021regularization}     & Industrial metal (pump) forming anomaly detection      & Manufacturing process monitoring  & Parameter transfer                             & Inductive     & LSTM                   & Single  &  \checkmark           &  \checkmark                 \\
\hline
\cite{parkMENDELTimeSeries}     & Industrial control systems anomaly detection      & Manufacturing process monitoring  & Parameter transfer                             & Inductive     & Auto-encoder                  & Single  &  \checkmark           &  \checkmark                 \\

\hline
\cite{panjapornpon2023explainable}   & Petrochemical production process anomaly detection            & Energy saving                     & Parameter transfer                             & Inductive     & LSTM, CNN        & Single  &  \checkmark                      &  \checkmark                 \\
\hline
\cite{liang2018transfer}             & Electricity consumption anomaly detection       & Energy saving                 & Parameter transfer                             & Inductive     & FCN                    & Single   &   \xmark                             &  \checkmark                  \\
\hline
\cite{copiaco2023innovative}          & Building’s energy consumption anomaly detection                       & Energy saving                                              & Parameter transfer                             & Inductive     & AlexNet-40            & Single   & \checkmark                          &  \checkmark                     \\
\hline
\cite{xu2021anomaly}                   & Power consumption anomaly detection             & Energy saving                                              & Mapping transfer              & Transductive & DAN      & Single  &  \checkmark               & \xmark               \\
\hline
\cite{simoneAnalysisMachineLearninga}   & Building’s power consumption  anomaly detection                       & Energy saving                                              & Parameter transfer                             & Inductive     & LSTM      & Single  &  \checkmark           &  \checkmark   \\
\hline
\cite{xiong2018application}             & Aircraft flight anomaly detection                                    & Infrastructure facilities monitoring                       & Parameter transfer                             & Inductive     & LSTM        & Single            &  \checkmark                         & \checkmark                     \\
\hline
\cite{pan2023transfer}                  & Anomaly identification for bridge groups            & Infrastructure facilities monitoring                       & Parameter transfer                             & Inductive     & CNN                    & Single  &  \checkmark             & \checkmark                  \\
\hline
\cite{dhillon2020towards}                & Network intrusion detection                         & Infrastructure facilities monitoring                       & Parameter transfer                             & Inductive     & CNN, LSTM               & Single  &  \checkmark            & \checkmark                \\
\hline
\cite{weberDetectionBuildingOccupancy2020}                & Building occupation detection                         & Infrastructure facilities monitoring                       & Parameter transfer                             & Inductive     & CNN              & Single  &  \checkmark            & \checkmark                \\
\hline
\cite{sayedTimeseries2DImages2023}                & Building occupation detection                         & Infrastructure facilities monitoring                       & Parameter transfer                             & Inductive     & CNN              & Single  &  \checkmark            & \checkmark                \\
\specialrule{0.6pt}{0pt}{0pt} 

\end{tabular}
\end{sidewaystable*} 





\subsection{Manufacturing process monitoring}
Manufacturing process monitoring is crucial to ensure high-quality products and low rejection rates.
For example, in injection molding machines, sensors are installed to detect molding conditions in the cavities, such as cavity pressure and temperature. These signals are used to analyze particularly the mold filling and solidification process for each produced part. Such cyclic processing data can also be seen in metal machining (cutting force signal) or joining of parts (joining force signal). 
Currently, parameter transfer is predominantly used for manufacturing processes
\cite{lockner2021induced,Tercan2018bridging,tercan2019industrial,lockner2022transfer,maschler2021towards,abdallah2021anomaly, abdallah2023anomaly,hsieh2019unsupervised,maschler2021regularization}.

Park \textit{et al.} propose a transfer learning technique to detect time series anomalies for different industrial control systems \cite{parkMENDELTimeSeries}. First, they apply principal components analysis to reduce the dimension of source and target data. A DNN model is then trained on the compressed source data, and after a reasonable mapping algorithm is adopted to map the features of source to target domain, the pre-trained model is further trained on the target data. The model achieves good performance even when a model is retrained with only a proportion of target data. For the experiments, they only test on two comparatively larger datasets and fail to show that the transferred model performs better than the one without retraining for one dataset. Further investigation needs to explain the negative transfer. Additionally, even when they only take a small proportion of target data for transfer learning purposes, the sample size still exceeds $5000$, which exceeds most industrial applications.
Abdallah \textit{et al.} apply parameter transfer to monitor the operation status of manufacturing testbeds with vibration sensor data \cite{abdallah2021anomaly, abdallah2023anomaly}. 
Hsieh \textit{et al.} transfer knowledge across three chambers in a production line to detect anomalous time series data \cite{hsieh2019unsupervised}. Results show reduced training time and improved detection accuracy through transfer learning. 
In injection molding, parameter transfer is applied to transfer the knowledge from one or more source domains to solve tasks in a target domain \cite{lockner2021induced,tercan2019industrial,lockner2022transfer}. 
Specifically, they employ simple fully connected neural networks and transfer knowledge from one product to another by freezing the first few layers and fine-tuning only the last few layers. Instead of evaluating the time series data directly from sensors, they represent the industrial process by the parameters of the machine settings. However, they can still provide useful insight for the case of the time-series data.
Tercan \textit{et al.} build a bridge between simulated data and real data using parameter transfer in injection molding \cite{Tercan2018bridging}. Here, a fully connected neural network is trained on simulated data and then partially or fully reused to further train on real data. Results show that the transferred model performs better than a network trained from scratch on real experimental data. In manufacturing processes, a simulation model/process hence can play a critical role, but deeper analysis is needed to further understand and reduce the gap between simulation and real data.
Additionally, Lockner \textit{et al.} explore the impact of transfer learning with varying amounts of source data and assess how performance is influenced by different configurations of frozen layers \cite{lockner2022transfer}. 
Maschler \textit{et al.} compare different DNNs for anomaly detection tasks on metal forming datasets \cite{maschler2021towards}. 
Further, they propose a deep transfer learning framework aiming to transfer knowledge between tasks. However, the proposed architecture is not validated by experiments. 
Later, Maschler \textit{et al.} apply continuous learning on the same dataset by transferring knowledge from several source tasks to a target task to train a deep
learning algorithm capable of solving both source and target tasks \cite{maschler2021regularization}. Specifically, they use regularization approaches using altered loss functions to solve related tasks that appear best suited.


\subsection{Predictive maintenance}
Predictive maintenance aims to predict the necessity of maintenance before production is negatively impacted by a failure. Tasks involve monitoring equipment to anticipate maintenance requirements (i.e., predict probable future failure) to optimize maintenance schedules \cite{serradilla2022deep}. Time series anomaly detection is often used in respective systems to identify abnormal behaviors in operation that may indicate the need for maintenance, such as increasing noise, vibrations, etc.

Mao \textit{et al.} use mapping transfer with a Sparse Auto-Encoder (SAE) for motor vibration anomaly detection \cite{mao2020robust}. A transformation from the source and target data to a common latent feature space is learned with MMD loss to make the feature distribution of two domains as identical as possible. 
Similarly, Wen \textit{et al.} also use mapping transfer with an SAE architecture for fault detection of rotation bearings, using an MMD regularizer to extract a common feature representation \cite{wen2017new}. 
Subsequently, they propose a new MU-Net architecture to detect multivariate time series anomalies \cite{wen2019time}. First, they pre-train a U-Net \cite{Ronneberger2015UNet} on a large time series dataset for an anomaly detection task.
Then, they propose a new model MU-Net, which is built upon U-Net. In MU-Net, each channel can leverage a pre-trained U-Net through fine-tuning to transfer knowledge for multivariate time series anomaly detection.

In a different application, parameter transfer is used to predict the remaining useful life for tools in manufacturing \cite{8540073}. An SAE network is first trained to predict the remaining useful life of a cutting tool on retrospectively acquired data in an offline process. The trained network is then transferred to production with a new tool for online remaining useful life prediction.
The result shows that transfer-learning based hybrid deep learning significantly reduces the training time and is highly suitable for real-time industrial fault diagnosis/prediction in various environments.
Similarly, parameter transfer is implemented to reduce the gap between different industrial environments \cite{zabin2022hybrid,xu2019digital}. 
Xu \textit{et al.} use a stacked SAE to extract general features from source data and a digital-twin-assisted fault diagnosis approach is presented to transfer knowledge from virtual space to physical space for real-time use \cite{xu2019digital}.
Here, a DNN model is first fully trained in virtual space and then migrated to the physical space using parameter transfer for real-time use. 

The above-mentioned literature proves that deep transfer learning is a research field that could simplify the life cycle of predictive maintenance systems and facilitate DNN model reusability by reducing the required data and training time, helping adapt them to solve similar tasks. 


\subsection{Energy management}
Energy management deals with systems that detect abnormal excessive consumption caused by end-users' unusual behavior or malfunction of faulty devices or systems \cite{copiaco2023innovative}. The goal is to develop automatic, quick-responding, accurate, and reliable fault detection to save energy and build environmentally friendly systems. Energy anomaly detection systems monitor data during energy generation, transmission, and utilization, to ensure normal energy consumption. 

Xu \textit{et al.} design a cluster-based deep adaptation layer to improve a deep adaptation network, effectively reducing the mismatch in transfer learning of spinning power consumption anomaly detection \cite{xu2021anomaly}.
The basic architecture consists of five convolutional layers and three fully connected layers. The weight parameters of the convolutional layers are shared between source and target domains. The cluster-based deep adaptation layer is designed across the feature layers of two networks to cluster feature representations of the source and target domains respectively. The proposed method shows superiority over fine-tuning and DAN because the adaptation layer can minimize the distance between the nearest neighbor clusters across the source and target domains to match the most similar distribution of feature representations. It is important to note that the anomalies are defined and tagged by human experts as different types, thus the problem becomes a classification task. However, in real-world industrial applications, it's almost impossible to enumerate unknown anomaly types because of the highly dynamic environment.
Liang \textit{et al.} successfully build an electricity consumption time series anomaly detection method in aluminum extrusion \cite{liang2018transfer}. Parameter transfer is applied to transfer domain knowledge from another data-sufficient domain. 
First, they train on sufficient extruding machine data in an unsupervised way and then use only a few data samples from different extruding machines to adapt the model by transfer learning. It is important to note that when the target data is already sufficient, transferring knowledge can be detrimental as it can decrease prediction accuracy on the final task.
Copiaco \textit{et al.} aims to detect anomalies for building energy consumption via transfer learning from pre-trained CNN models \cite{copiacoExploringDeepTimeSeries2022}. First, they convert 1D time series signals to 2D image representations. These serve as inputs for pre-trained vision models to capture inherent spatially invariant features. In the end, a SVM is applied to classify anomaly types. The SVM classifier obtained optimal results when operating upon a pre-trained ALexNet model with normalized grayscale graphical representations. However, a deeper discussion regarding the effect of the different pre-trained models is not presented. Additionally, converting 1D time series to 2D images by creating a matrix representation of the sensor readings may lead to information loss during the transformation process, which should be further investigated.

\subsection{Infrastructure facilities monitoring}
Infrastructure facilities monitoring refers to monitoring and maintaining the conditions of infrastructure facilities, such as bridges, buildings\cite{sayedDeepTransferLearning2022}, and networks. This can include detecting potential problems or failures. The goal is to minimize the impact of failures on the public or the environment.
This application commonly uses parameter transfer to transfer knowledge from facility to facility to take advantage of similar data and tasks. 

Dhillon \textit{et al.} present a parameter transfer approach towards building a network intrusion detection system based on CNN and LSTM \cite{dhillon2020towards}. Specifically, They extract and learn patterns by mapping the input data into a lower dimensional representation by convolutional layers. Then, they employ the LSTM layer to enhance learning and recognizing patterns across time. In the end, a fully connected layer is used as a classifier to predict normal and malicious data. To do the parameter transfer, they reuse the model architecture and freeze most weight parameters for the target domain so that they do not need a large training dataset to retrain the model. However, they do not mention implementation details, like which layers are frozen in the transfer learning stage. Observing how transfer learning performs with different frozen layers would be interesting. 
Pan \textit{et al.} apply parameter transfer to fully use the similarity of the anomalous patterns across different bridges \cite{pan2023transfer}. They train a CNN model on one bridge data, then transfer the knowledge obtained by the CNN model to a small part of the target data. They update the last three fully connected layers while keeping the convolutional layers intact.  The experimental results show transfer learning achieves higher accuracy anomaly detection across bridges. 
Weber \textit{et al.} takes advantage of simulation data by training on synthetic environmental data, then fine-tunes the pre-trained model and transfers the knowledge from simulation data for real-time online building occupancy detection \cite{weberDetectionBuildingOccupancy2020}. Although the results show the effectiveness of transfer learning, the availability, and reliability of the simulation data for other industrial applications is still an open issue. 
Sayed \textit{et al.} adapt parameter transfer using pre-trained CNN models, such as AlexNet and GoogLeNet, pre-trained on ImageNet \cite{sayedTimeseries2DImages2023}. The pre-trained model is then further used for downstream tasks. The results show the pre-trained models outperform their customed CNN model, which is not pre-trained. However, it is important to note that the transfer effect may not be entirely convincing due to the fact that the customer CNN model is not pre-trained on the same dataset.

\subsection{Application-independent considerations}
\add{
Data scarcity, as well as domain shift, stand out as the two main common problems independent of the industrial field of application. The same problems have originally prompted the use of transfer learning in general, and respectively, general techniques are applied widely across domains. Regarding data scarcity, this un-surprisingly involves leveraging pre-trained models as a starting point for further training. Regarding domain shift, mapping transfer and parameter transfer are the most often-used approaches. Unlike parameter transfer, mapping transfer incorporates the source and target data in the training process. Instance transfer and domain-adversarial transfer learning were not employed in the surveyed literature -- researchers seem to not see huge value in these methods for the surveyed fields.

Another common aspect across time series anomaly detection applications is the choices of model architecture to facilitate the training process by capturing temporal dependencies and recognizing patterns over varying time scales: Favourite architectures include CNNs, LSTMs, and auto-encoders that have sets of assumptions (inductive biases) about the data they analyze that make them excel in understanding the sequential nature of time series data. 
CNNs assume local (in time) connectivity, stationary statistics, and hierarchical structure, and induce certain translation-invariance.
LSTMs are still given preference in many applications over the more modern deep learning architecture of choice for sequence learning, the transformer. The reason is their stronger inductive bias, leading to less data (and compute) hunger.
Both CNN and LSTM networks can be built as e.g. classifiers, but also auto-encoders. These latter architectures have the advantage of learning low-dimensional representations of the high-dimensional time series with minimal loss of signal in an unsupervised way. The analysis of the data in the low-dimensional latent space facilitates anomaly detection. The concrete choice of architecture does not depend on the field of application but on the data and task, and thus the most suitable inductive bias.

As an interim conclusion, the most striking application-independent finding is that across the surveyed literature, predominantly simple, tried-and-tested design patterns for transfer learning are used in industry. The field of deep transfer learning would offer a much wider variety of approaches.}

\section{Discussion}
\label{sec_discussion}
\subsection{Potential}
The automation of industrial process monitoring stands as a transformative step toward increasing efficiency and optimizing quality. While standard deep learning training is sufficient in discerning intricate patterns from vast datasets, its application in the dynamic industrial landscape is not without challenges. Chief among them is the impracticality of continuously obtaining large-scale labeled data to train models afresh for every nuanced variation in processes. Deep transfer learning has shown promise with its adaptive capabilities. By mitigating the need for extensive labeled data and eliminating the necessity to train models from scratch for every distinct setup.
However, adopting deep transfer learning beyond simple parameter transfer is still a challenge.


\subsection{Challenges}
\paragraph{Domain shift}
Different from the i.i.d assumption in most machine learning problem settings, many industrial processes suffer from substantial domain shift due to dynamic changes in industrial settings, e.g., change of products or measuring sensors. 
Domain shift lies at the heart of the deep transfer learning problem. Particularly, the dynamic changes in many industrial processes, up to an apparent dissimilarity of source and target data, make the transfer learning task particularly challenging.
Here we list the key challenges associated with domain shift: 
\begin{itemize}
    \setlength\itemsep{0.5em}
    \item \textit{Covariate shift} occurs when the marginal distribution of the features changes from source domain to target domain. The distribution mismatch poses challenges in transferring the knowledge from source to target domain.
    \item \textit{Concept shift} refers to the changes in the relationships between features and labels. The relationship can change from source domain to target domain, leading to bias and error in the model. 
    \item \textit{Label shift} refers to the label distribution in the target domain that can be different from the source domain, whether the marginal distribution changes or not. 
\end{itemize}

\paragraph{Label availability and reliability}
Deep transfer learning is built upon deep learning, which usually requires a large amount of labeled data, the more data a model has available for training, the better it can generalize to new examples.
In real-world industrial time series anomaly detection tasks, collecting data is probably easy, but collecting labels is much more expensive and time-consuming, sometimes prohibitively so, leading to the unavailability of sufficient labeled data.
Self-supervised learning can be used to re-label a large amount of unlabeled data and thus 
anomaly detection models usually need to learn in an unsupervised or semi-supervised mode \cite{goldstein2016comparative}. In industrial cases, another significant concern is to ensure the data quality. Due to the high cost associated with obtaining reliable and precise labels, usually self-supervised learning is applied to create pseudo labels or relabel the unlabeled data, thus facilitating the transfer learning process. Additionally, a data-centric process with humans in the loop can be involved in improving label reliability. However,  unreliable labels can still affect the transfer learning training process.

\paragraph{Missing relevant data information}
Missing relevant data poses significant challenges for transfer learning since it can affect the model's ability to generalize and transfer knowledge from source to target domain. 

\begin{itemize}
    \setlength\itemsep{0.5em}
    \item \textsl{Imbalanced data}: Even if the labels can be collected, anomalies can be extremely rare by design, which poses the risk of training with extremely imbalanced data. A practical problem for anomaly detection in industry is the extremely imbalanced data distribution, in which normal samples dominate in data and abnormal samples only share a small percentage in the whole dataset. 
Prior research has proven that the effect of class imbalance on classification performance by using deep learning is detrimental \cite{buda2018systematic}. However, most research studies still ignore such problems, which can result in poor performance regarding the minority class, i.e., abnormal data are misclassified as normal.
    \item \textsl{Information loss}: missing data can lead to lost important features. For example, some information that has a significant effect on the process from case to case is not even recorded or is too complex to record (i.e. part geometry, machine geometry, or environmental conditions in injection molding processes).
\end{itemize}

Various approaches have been developed to address these challenges to reduce the domain gap between the source and target domains, aiming at mitigating domain shift. 
These techniques involve domain generalization, contrastive learning, and adversarial examples.
The domain shift problem is far from being solved. To tackle this problem, transfer learning requires a deep understanding of the target data's characteristics and appropriate transferable strategies to effectively bridge the gap between the source and target domains.

\paragraph{Effectiveness of deep transfer learning}
The general effectiveness of deep transfer learning is limited by the difficulty of determining which knowledge or to what extent the knowledge should be transferred from source to target task. Unlike natural language processing, pre-training a language model on a large corpus of text data can help the model learn the statistical patterns and semantic and syntactic representations of words and sentences, which can be used for new natural language processing tasks with a few or even without data. Due to data privacy, large available public datasets usually do not exist for industrial time series, or they can not be used because of a large domain gap between different datasets and tasks. 
In this case, transferring all of the knowledge may not be beneficial, as it may be irrelevant.
In the worst case, this can lead to negative transfer \cite{pan2010survey, smith2001transfer}, in which the extracted knowledge harms the new task-learning.
This requires assessing how source and target tasks are related, carefully selecting the knowledge to be transferred, and selecting the proper means to implement this transfer. 
Glorot \textit{et al.} attempt to analyze and quantify the gained knowledge from source to target domain \cite{glorotDomainAdaptationLargeScale2011}. For example, they define transfer error, transfer loss, transfer ratio, and in-domain ratio, which provide metrics to interpret the transferring performance.


\subsection{Directions for anomaly detection solution design}
\paragraph{Data preprocessing}
How data preprocessing should be conducted is an open question. 
For industrial applications, some researchers contend that using raw time series data directly as input for training may not be the most efficient. Hence, they propose deriving or selecting features from time series data by statistical methods or human experience. This can significantly decrease the complexity of the dataset. On the other hand, this crops a lot of potentially useful information, e.g., the time series trend. 
To reduce the dimensionality, some researchers use machine parameters as features in the manufacturing process instead of the processing data collected by sensors \cite{Tercan2018bridging,lockner2021induced, tercan2019industrial,lockner2022transfer}. 
Others try different transformations of raw time series data, a common way being to transform 1D time series data to 2D image data \cite{liao2021manufacturing,wang2021anomaly,zabin2022hybrid} or transforming time domain signals otherwise into the frequency domain \cite{7738816}.
However, as large-scale computation power and storage become cheaper and more accessible, it is becoming increasingly common to use deep learning techniques to process time series data directly \cite{xu2019digital, maschler2021towards}.

\paragraph{Data augmentation by generative AI}
Data augmentation is useful for deep learning models because it can help to prevent overfitting. For deep transfer learning, when a model becomes too closely adapted to the specifics of the source domain, it may not be able to generalize well to some examples in the task domain.
One important technique is to acquire effective synthetic data, e.g., using a simulation process or model to explore potential anomalous conditions by simulating industrial processes under parameters that cannot yet be experienced in the real world. High fidelity and reliable simulation data can provide training data at low cost and mitigate the problem of insufficient samples for deep transfer learning \cite{xu2019digital}.
Another way to generate effective synthetic data is to use generative models, such as GANs. 
GANs are only trained on normal data to generate indistinguishable normal samples so that abnormal samples can be distinguished during the testing stage of the overarching anomaly detection system, as they deviate from the normal data distribution \cite{8853246}. To increase the number of anomalous samples and thus the robustness of the anomaly detection model, the technique of adversarial perturbation known from computer vision \cite{goodfellow2014explaining} can be used. 

\paragraph{Dealing with data imbalance}
DNNs perform well when they are trained on balanced datasets. However, in practice, it is difficult to get sufficient anomalous data for anomaly detection tasks. For example, the manufacturing process is usually in a healthy state due to the pre-designed and optimized operation. 
Several ways exist to address the imbalanced dataset for time series anomaly detection. 
One way is to oversample the minority class, e.g., by randomly replicating samples from the minority class to equalize the number of samples from each class in each batch. The Synthetic Minority Over-sampling Technique is an advanced method that creates synthetic samples to force the decision region of the minority class to become more general \cite{chawla2002smote}. This technique is widely used in anomaly detection tasks in industry \cite{ijaz2018hybrid, mokhtari2021machine}. 
Apart from oversampling, resampling strategies are frequently used to assign a higher probability to abnormal samples and evenly select the same amount of samples from both classes in each batch. Moreover, a weighted loss can be implemented to balance the loss between the abnormal and normal class in supervised anomaly detection \cite{buda2018systematic}.

\subsection{Directions for deep transfer learning implementation}
\paragraph{When shall deep transfer learning be used?}
(1)  \change[]{Limited data availability: If the data available for a specific task is limited, pre-training on related source data can learn general features that can be transferred to the specific learning task in the target domain.}
{Limited data availability: It poses a significant challenge in machine learning, particularly when aiming to train models for specific tasks. Pre-training a model on a larger or more diverse dataset, even if unrelated to the specific task at hand, enables the acquisition of generalizable features and representations. These generalized features, learned from a broader context, can then be effectively transferred to analyze the target domain with limited data. This can effectively provide a practical solution to the challenges posed by data scarcity.
}
(2)  Similar domains: Deep transfer learning is well suited when tackling source and target domains with a high degree of similarity. 
\add[]{In such instances, knowledge can be derived either from a model pre-trained on a similar dataset or one trained on both source and target data.
In both cases, the model can efficiently transfer relevant features and representations within the domain, facilitating a more robust adaptation to the target dataset, and ultimately optimizing the model's ability to discern and detect patterns within the target domain.}
(3)  \change[]{Limited resources (time and compute): In cases where resources are constrained, it is recommended to employ parameter transfer, especially if a pre-trained model is readily available.}
{Limited resources (encompassing both time and computational power): When faced with resource constraints, it is recommended to employ parameter transfer, especially if a pre-trained model is readily available.
As described in} \cite{torrey2009transferlearning}, \add[]{the transfer might improve learning in three distinct ways: (a) a higher performance at the very beginning of learning, (b) a steeper slope in the learning curve, or (c) a higher asymptotic performance.
Parameter transfer leverages the learned parameters and weights of a pre-trained model, often trained on a larger dataset. By doing so, the resource-intensive process of training a model from the ground up is circumvented, and the computational burden is significantly alleviated. }

\paragraph{When not to use deep transfer learning?}
(1)  Irrelevant data: If the target data is vastly different from the source data, deep transfer learning may not be appropriate, sometimes even leading to negative transfer. For example, if one wants to train a model for natural language processing on a new dataset, using a pre-trained model that has been trained on image data may not yield meaningful results. This is due to the vast dissimilarity in data modalities and features between images and text.
\add[]{(2) Task-specific models:
In scenarios where the target task is well-defined and specific, and pre-trained models do not align closely with the task requirements, it is usually more effective to build a task-specific model from scratch.}
(3)  High domain shift: If there is a large difference between the source and the target domain, deep transfer learning may not be effective. This can happen when the data distributions, features, or labels are vastly different.
(4)  Abundance of labeled data available for target task: If there are enough data for the new task, it may be more effective to train a model from scratch \cite{liang2018transfer}.

\paragraph{What model architecture to choose?} 
We recommend choosing the model architecture mainly based on data size and label availability, starting from a relatively small network and moving gradually to more complex DNNs. 
CNNs also effectively extract time series features \cite{yao2022model, zgraggen2021transfer}. For semi-supervised settings, CNN-based auto-encoders are trained to reconstruct the original data \cite{serradilla2021adaptable}. 
It is important to effectively capture the temporal dependencies and extract features of time series data.
LSTMs are extensively employed for this purpose, as they excel in detecting temporal dependencies in time series data \cite{zabin2022hybrid, panjapornpon2023explainable, xiong2018application}. 

\add[]{
Exploring hybrid architectures that combine the advantages of CNNs, RNNs, and LSTMs for tasks involving both spatial and temporal dependencies can be beneficial.} Cao \textit{et al.} \add[]{propose a multi-head CNN–RNN architecture for multi-time series anomaly detection} \cite{canizoMultiheadCNNRNN2019}. \add[]{A CNN is used to extract meaningful features from raw data and then an RNN is applied to learn temporal patterns simultaneously. Similarly, Dhillon \textit{et al.} utilize LSTM layers to model the time series signals after obtaining the features from a CNN.
An alternative way to benefit from different models is the use of ensemble approaches, combining the strengths of different model architectures to enhance performance, especially in situations where the target task requires capturing diverse features.
In the future, we expect more applications to use transformer-based approaches as pre-trained models become available and public datasets get open-sourced.}



\paragraph{Beyond transfer learning}
Foundation models like SAM \cite{segmentationAnything2023} or others, using for example transformer architectures \cite{vaswani2017attention} or diffusion models \cite{Rombach2022}, demonstrate
emerging properties such as in-context learning \cite{NEURIPS2020_1457c0d6} and complex cross-modality conditioning.
This is achieved by training complex and often auto-regressive models with massive amounts of data,
although the precise mechanisms that lead to this are not well understood.
Some of those models generalize to new settings and tasks, without an explicit element of transfer learning.
Thus, the application of foundation models in industrial time series analysis has the potential to
reduce and eventually eliminate the need to explicitly account for changes in the domain within the modeling, by instead having the foundation model provide the transfer capability (see examples in Sec. \ref{sec_conclusions}-b).
To not only detect anomalies but also identify failure modes, analyze root causes, and elicit an appropriate intervention,
AI systems must implicitly or explicitly model causal relations.
Counterfactual inference incorporates causal relations between observations and interventions, which
allows predictions of outcomes never seen during training \cite{estimating2023}.

Another aspect of deep learning implementations is the limited computing power of hardware platforms, such as embedded systems in industry. Sensor data are typically acquired using resource-constrained edge processing devices that struggle with computationally intensive tasks, especially when training a DNN model. 
Federated learning stands out as a leading solution, with its ability to utilize data while preserving privacy \cite{wang2018edge, amiri2020machine}. The technology enables a more collaborative approach to ML while preserving user privacy by storing data decentralized on distributed devices rather than on a central server. Combining deep transfer learning with federated learning is a promising and powerful combination in the abovementioned industrial applications.

\section{Conclusions}
\label{sec_conclusions}


In this survey, we presented a comprehensive overview of deep transfer learning by defining transfer learning problem settings and categorizing the state-of-the-art deep transfer learning approaches based on the surveyed papers.
Then, we review and emphasize on investigating deep transfer learning approaches for time series anomaly detection in different industrial settings.
Equipped with this foundation, we selected representative examples of the landscape of fielded applications to provide practitioners with a guide to the field and possibilities of industrial time series anomaly detection.

The main finding of this survey is that only a limited variety of deep transfer learning methods
are employed in anomaly detection in industrial time series analysis -- mainly simple ones.
Almost all applications employ parameter transfer, arguably the most straightforward transfer approach. In its simplest implementation, it only involves fine-tuning a pre-trained model.
Accordingly, the employed network architectures are simple, none of the reviewed research papers used advanced DNN building blocks like Transformer,
which are common in computer vision and language modeling.
We expect this type of architecture with suitable modifications and/or pre-trained parameters to spread to more niche fields.
Despite this, the survey suggests that deep transfer learning approaches have huge potential and promise for solving more complex and dynamic anomaly detection tasks in industry. As the field is still in an early stage, more R\&D is expected to fully realize the potential of deep transfer learning in increasingly complex settings.

In the end, we highlight the importance of considering feasibility, reliability, explainability, and real-time data stream when designing a transfer learning system for time series anomaly detection.
After carefully discussing open challenges, we gave practical directions for time series anomaly detection solution design and deep transfer learning implementation.
In our view, the following directions hold the greatest potential for future work: 

\paragraph{Automatic selection of transferable features \cite{long2015learning} } It refers to methods for selecting and transferring only the relevant knowledge for the new tasks from the base model. This could involve the use of techniques such as selective fine-tuning and distillation to identify the most important features learned from source domains \cite{yosinski2014transferable, Ge_2017_CVPR}. 
\paragraph{Investing into advanced deep transfer learning schemes and DNN models}
The conceptionally simplest parameter transfer approach has the advantage of being readily applicable by interdisciplinary teams without ML research experience. However, it seems promising to invest in testing more sophisticated deep transfer learning approaches according to different use cases, such as mapping transfer, adversarial transfer, etc. The same applies to testing diverse DNN models besides straightforward ones. \add[]{Recently, large models have been used in time series anomaly detection. }
\add[]{For example, Xu \textit{et al.} propose the Anomaly Transformer with a new anomaly-attention mechanism to compute the association discrepancy} \cite{xuAnomalyTransformerTime2021}.
\add[]{A minimax strategy is devised to amplify the normal-abnormal distinguishability of the association discrepancy. 
On the other hand, Pintilie \textit{et al.} leverage diffusion models for multivariate time series anomaly detection} \cite{pintilie2023}. \add[]{They train two diffusion-based models that outperform strong transformer-based methods on synthetic datasets and are competitive on real-world data. Additionally, their DiffusionAE model is more robust to different levels and the number of anomaly types. 
These large models have proven to be effective and advantageous given certain data and tasks. 
It's important to note that their effectiveness also depends on the characteristics of the time series data and the requirements of the anomaly detection task.
Additionally, model computational efficiency and interpretability should be considered, especially in real-time or resource-constrained industrial applications. }
\paragraph{Data-centric approach to real-time anomaly detection} The data-centric approach focuses on improving ML models by ensuring high-quality labeled data \cite{stadelmann2022data} using techniques such as re-labeling, re-weighting, or data augmentation \cite{luley2023}. Currently, a human-in-the-loop solution is still needed. Frameworks have been proposed to assist annotators with graph-based algorithms such as nearest neighbor graphs \cite{bai2021self}, decision trees \cite{liu2021autodc}, or factor graphs \cite{kang2022finding}. Although these methods have proven to be effective, a more automated process is a goal for future research. 
\paragraph{Leveraging generative AI} 
Generative models like GANs and diffusion models can generate synthetic time series data, making them valuable for data augmentation.
Augmenting the original data with synthetic samples can enhance the deep learning models' robustness, especially in real-world applications where target data are limited. These models can also be leveraged to examine anomalies and generate anomalies to help alleviate the imbalance within the data \cite{salemAnomalyGenerationUsing2018}. 

\paragraph{Integration with other ML methods} 
To develop robust AI solutions for time series anomaly detection in the industry, relying solely on transfer learning is insufficient. Future strategies should integrate other ML approaches, including continuous learning, meta-learning, and federated learning.










\bibliographystyle{unsrt}
\bibliography{IEEE_Acess_final_after_proof_arxiv}

\begin{thebibliography}{100}

\bibitem{kagermann2011industrie}
Henning Kagermann, Wolf-Dieter Lukas, and Wolfgang Wahlster.
\newblock Industrie 4.0: {Mit} dem internet der dinge auf dem weg zur 4. industriellen revolution.
\newblock {\em VDI Nachrichten}, 13(1):2--3, 2011.

\bibitem{Roblek2016}
Vasja Roblek, Maja Meško, and Alojz Krapež.
\newblock A complex view of industry 4.0.
\newblock {\em SAGE Open}, 6(2), April 2016.

\bibitem{Wang2018}
Lihui Wang, Martin Törngren, and Mauro Onori.
\newblock Current status and advancement of cyber-physical systems in manufacturing.
\newblock {\em J. Manuf. Syst.}, 37:517--527, October 2015.

\bibitem{Jeschke2017}
Sabina Jeschke, Christian Brecher, Tobias Meisen, Denis Özdemir, and Tim Eschert.
\newblock Industrial internet of things and cyber manufacturing systems.
\newblock In {\em Industrial {Internet} of {Things}: {Cybermanufacturing} {Systems}}, pages 3--19. Springer, 2017.

\bibitem{Dalenogare2018}
Lucas~Santos Dalenogare, Guilherme~Brittes Benitez, Néstor~Fabián Ayala, and Alejandro~Germán Frank.
\newblock The expected contribution of {Industry} 4.0 technologies for industrial performance.
\newblock {\em Int. J. Prod. Econ.}, 204:383--394, October 2018.

\bibitem{Kagermann2017}
Henning Kagermann.
\newblock Chancen von {Industrie} 4.0 nutzen.
\newblock In {\em Handbuch {Industrie} 4.0 {Bd}.4: {Allgemeine} {Grundlagen}}, pages 237--248. Springer, Berlin, Heidelberg, 2017.

\bibitem{9381850}
S.~M. Abu~Adnan Abir, Adnan Anwar, Jinho Choi, and A.~S.~M. Kayes.
\newblock {IoT}-{Enabled} smart energy grid: {Applications} and challenges.
\newblock {\em IEEE Access}, 9:50961--50981, 2021.

\bibitem{parkReviewFaultDetection2020}
You-Jin Park, Shu-Kai~S. Fan, and Chia-Yu Hsu.
\newblock A review on fault detection and process diagnostics in industrial processes.
\newblock {\em Processes}, 8(9):1123, September 2020.

\bibitem{liao2021manufacturing}
Yabin Liao, Ihab Ragai, Ziyun Huang, and Scott Kerner.
\newblock Manufacturing process monitoring using time-frequency representation and transfer learning of deep neural networks.
\newblock {\em J. Manuf. Processes}, 68:231--248, August 2021.

\bibitem{lockner2021induced}
Yannik Lockner and Christian Hopmann.
\newblock Induced network-based transfer learning in injection molding for process modelling and optimization with artificial neural networks.
\newblock {\em IJAMT}, 112(11):3501--3513, February 2021.

\bibitem{maschler2021towards}
Benjamin Maschler, Tim Knodel, and Michael Weyrich.
\newblock Towards deep industrial transfer learning for anomaly detection on time series data.
\newblock In {\em Proc. 26th {IEEE} {ETFA}}, pages 01--08, September 2021.

\bibitem{wen2019time}
Tailai Wen and Roy Keyes.
\newblock Time series anomaly detection using convolutional neural networks and transfer learning, May 2019.
\newblock arXiv:1905.13628.

\bibitem{xu2019digital}
Yan Xu, Yanming Sun, Xiaolong Liu, and Yonghua Zheng.
\newblock A digital-twin-assisted fault diagnosis using deep transfer learning.
\newblock {\em IEEE Access}, 7:19990--19999, 2019.

\bibitem{mao2020robust}
Wentao Mao, Di~Zhang, Siyu Tian, and Jiamei Tang.
\newblock Robust detection of bearing early fault based on deep transfer learning.
\newblock {\em Electronics}, 9(2):323, February 2020.

\bibitem{wang2021anomaly}
Weiping Wang, Zhaorong Wang, Zhanfan Zhou, Haixia Deng, Weiliang Zhao, Chunyang Wang, and Yongzhen Guo.
\newblock Anomaly detection of industrial control systems based on transfer learning.
\newblock {\em Tsinghua Sci. Technol.}, 26(6):821--832, December 2021.

\bibitem{canizoMultiheadCNNRNN2019}
Mikel Canizo, Isaac Triguero, Angel Conde, and Enrique Onieva.
\newblock Multi-head {CNN}–{RNN} for multi-time series anomaly detection: {An} industrial case study.
\newblock {\em Neurocomputing}, 363:246--260, October 2019.

\bibitem{weberDetectionBuildingOccupancy2020}
Manuel Weber, Christoph Doblander, and Peter Mandl.
\newblock Towards the detection of building occupancy with synthetic environmental data, October 2020.
\newblock arXiv:2010.04209.

\bibitem{sayedTimeseries2DImages2023}
Aya~Nabil Sayed, Yassine Himeur, and Faycal Bensaali.
\newblock From time-series to {2D} images for building occupancy prediction using deep transfer learning.
\newblock {\em EAAI}, 119:105786, March 2023.

\bibitem{yao2022model}
Y~Yao, D~Ge, J~Yu, and M~Xie.
\newblock Model-based deep transfer learning method to fault detection and diagnosis in nuclear power plants.
\newblock {\em Front. Energy Res.}, 10, 2022.

\bibitem{abdallah2021anomaly}
Mustafa Abdallah, Wo~Jae Lee, Nithin Raghunathan, Charilaos Mousoulis, John~W. Sutherland, and Saurabh Bagchi.
\newblock Anomaly detection through transfer learning in agriculture and manufacturing {IoT} systems, February 2021.
\newblock arXiv:2102.05814.

\bibitem{panjapornpon2023explainable}
Chanin Panjapornpon, Santi Bardeeniz, Mohamed~Azlan Hussain, and Patamawadee Chomchai.
\newblock Explainable deep transfer learning for energy efficiency prediction based on uncertainty detection and identification.
\newblock {\em Energy and AI}, 12:100224, April 2023.

\bibitem{dhillon2020towards}
Harsh Dhillon and Anwar Haque.
\newblock Towards network traffic monitoring using deep transfer learning.
\newblock In {\em Proc. {IEEE} 19th {TrustCom}}, pages 1089--1096, December 2020.

\bibitem{xiong2018application}
Peng Xiong, Yonxin Zhu, Zhanrui Sun, Zihao Cao, Menglin Wang, and Yu~Zheng.
\newblock Application of transfer learning in continuous time series for anomaly detection in commercial aircraft flight data.
\newblock In {\em Proc. {IEEE} {SmartCloud}}, pages 13--18, September 2018.

\bibitem{Bengio2014_Represen}
Yoshua Bengio, Aaron Courville, and Pascal Vincent.
\newblock Representation learning: {A} review and new perspectives.
\newblock {\em IEEE TPAMI}, 35(8):1798--1828, August 2013.

\bibitem{Maschler2022}
Benjamin Maschler, Hannes Vietz, Hasan Tercan, Christian Bitter, Tobias Meisen, and Michael Weyrich.
\newblock Insights and example use cases on industrial transfer learning.
\newblock {\em Procedia CIRP}, 107:511--516, 2022.

\bibitem{Maschler2021TLinIA}
Benjamin Maschler and Michael Weyrich.
\newblock Deep transfer learning for industrial automation: {A} review and discussion of new techniques for data-driven machine learning.
\newblock {\em IEEE Ind. Electron. Mag.}, 15(2):65--75, 2021.

\bibitem{zhuang2020comprehensive}
Fuzhen Zhuang, Zhiyuan Qi, Keyu Duan, Dongbo Xi, Yongchun Zhu, Hengshu Zhu, Hui Xiong, and Qing He.
\newblock A comprehensive survey on transfer learning.
\newblock {\em Proc. IEEE}, 109(1):43--76, January 2021.

\bibitem{pan2010survey}
Sinno~Jialin Pan and Qiang Yang.
\newblock A survey on transfer learning.
\newblock {\em IEEE Trans. Knowl. Data Eng.}, 22(10):1345--1359, October 2010.

\bibitem{tan2018survey}
Chuanqi Tan, Fuchun Sun, Tao Kong, Wenchang Zhang, Chao Yang, and Chunfang Liu.
\newblock A survey on deep transfer learning.
\newblock In {\em Proc. {ICANN} 2018}, pages 270--279, 2018.

\bibitem{yosinski2014transferable}
Jason Yosinski, Jeff Clune, Yoshua Bengio, and Hod Lipson.
\newblock How transferable are features in deep neural networks?
\newblock In {\em Proc. {NeurIPS}}, volume~27, 2014.

\bibitem{yuSurveyDeepTransfer2022}
Fuchao Yu, Xianchao Xiu, and Yunhui Li.
\newblock A survey on deep transfer learning and beyond.
\newblock {\em Mathematics}, 10(19):3619, January 2022.

\bibitem{choi2021deep}
Kukjin Choi, Jihun Yi, Changhwa Park, and Sungroh Yoon.
\newblock Deep learning for anomaly detection in time-series data: review, analysis, and guidelines.
\newblock {\em IEEE Access}, 9:120043--120065, 2021.

\bibitem{chandola2009anomaly}
Raghavendra Chalapathy and Sanjay Chawla.
\newblock Deep learning for anomaly detection: {A} survey, January 2019.
\newblock arXiv:1901.03407.

\bibitem{torrey2009transferlearning}
L.~Torrey and J.~Shavlik.
\newblock Transfer learning.
\newblock {\em Handbook of Research on Machine Learning Applications}, January 2009.

\bibitem{caruana1997multitask}
Rich Caruana.
\newblock Multitask learning.
\newblock {\em Mach. Learn.}, 28(1):41--75, 1997.

\bibitem{ruder2017overview}
Sebastian Ruder.
\newblock An overview of multi-task learning in deep neural networks, June 2017.
\newblock arXiv:1706.05098.

\bibitem{Kouw2019}
Wouter~M. Kouw and Marco Loog.
\newblock An introduction to domain adaptation and transfer learning, January 2019.
\newblock arXiv:1812.11806.

\bibitem{schmidhuber2015deep}
Jürgen Schmidhuber.
\newblock Deep learning in neural networks: {An} overview.
\newblock {\em Neural Networks}, 61:85--117, January 2015.

\bibitem{7738816}
Yanick Lukic, Carlo Vogt, Oliver Dürr, and Thilo Stadelmann.
\newblock Speaker identification and clustering using convolutional neural networks.
\newblock In {\em Proc. 26th {IEEE} {MLSP}}, pages 1--6, September 2016.

\bibitem{inbook111}
Thilo Stadelmann, Mohammadreza Amirian, Ismail Arabaci, Marek Arnold, Gilbert~François Duivesteijn, Ismail Elezi, Melanie Geiger, Stefan Lörwald, Benjamin~Bruno Meier, Katharina Rombach, and Lukas Tuggener.
\newblock Deep learning in the wild.
\newblock In {\em Proc. 8th {ANNPR}}, pages 17--38, 2018.

\bibitem{schmidhuber2022annotated}
Juergen Schmidhuber.
\newblock Annotated history of modern {AI} and deep learning, December 2022.
\newblock arXiv:2212.11279.

\bibitem{he2022instance}
Qi-Qiao He, Shirley Weng~In Siu, and Yain-Whar Si.
\newblock Instance-based deep transfer learning with attention for stock movement prediction.
\newblock {\em Appl. Intell.}, 53(6):6887--6908, July 2022.

\bibitem{amirian2021prepnet}
Mohammadreza Amirian, Javier~A. Montoya-Zegarra, Jonathan Gruss, Yves~D. Stebler, Ahmet~Selman Bozkir, Marco Calandri, Friedhelm Schwenker, and Thilo Stadelmann.
\newblock {PrepNet}: {A} convolutional auto-encoder to homogenize {CT} scans for cross-dataset medical image analysis.
\newblock In {\em Proc. 14th {CISP}-{BMEI}}, pages 1--7, October 2021.

\bibitem{wang2019instance}
Tianyang Wang, Jun Huan, and Michelle Zhu.
\newblock Instance-based deep transfer learning.
\newblock In {\em Proc. {IEEE} {WACV}}, pages 367--375, January 2019.

\bibitem{bommasani2021opportunities}
Rishi Bommasani, Drew~A. Hudson, Ehsan Adeli, Russ Altman, Simran Arora, Sydney von Arx, Michael~S. Bernstein, Jeannette Bohg, Antoine Bosselut, Emma Brunskill, Erik Brynjolfsson, Shyamal Buch, Dallas Card, Rodrigo Castellon, Niladri Chatterji, Annie Chen, Kathleen Creel, Jared~Quincy Davis, Dora Demszky, Chris Donahue, Moussa Doumbouya, Esin Durmus, Stefano Ermon, John Etchemendy, Kawin Ethayarajh, Li~Fei-Fei, Chelsea Finn, Trevor Gale, Lauren Gillespie, Karan Goel, Noah Goodman, Shelby Grossman, Neel Guha, Tatsunori Hashimoto, Peter Henderson, John Hewitt, Daniel~E. Ho, Jenny Hong, Kyle Hsu, Jing Huang, Thomas Icard, Saahil Jain, Dan Jurafsky, Pratyusha Kalluri, Siddharth Karamcheti, Geoff Keeling, Fereshte Khani, Omar Khattab, Pang~Wei Koh, Mark Krass, Ranjay Krishna, Rohith Kuditipudi, Ananya Kumar, Faisal Ladhak, Mina Lee, Tony Lee, Jure Leskovec, Isabelle Levent, Xiang~Lisa Li, Xuechen Li, Tengyu Ma, Ali Malik, Christopher~D. Manning, Suvir Mirchandani, Eric Mitchell, Zanele Munyikwa, Suraj Nair,
  Avanika Narayan, Deepak Narayanan, Ben Newman, Allen Nie, Juan~Carlos Niebles, Hamed Nilforoshan, Julian Nyarko, Giray Ogut, Laurel Orr, Isabel Papadimitriou, Joon~Sung Park, Chris Piech, Eva Portelance, Christopher Potts, Aditi Raghunathan, Rob Reich, Hongyu Ren, Frieda Rong, Yusuf Roohani, Camilo Ruiz, Jack Ryan, Christopher Ré, Dorsa Sadigh, Shiori Sagawa, Keshav Santhanam, Andy Shih, Krishnan Srinivasan, Alex Tamkin, Rohan Taori, Armin~W. Thomas, Florian Tramèr, Rose~E. Wang, William Wang, Bohan Wu, Jiajun Wu, Yuhuai Wu, Sang~Michael Xie, Michihiro Yasunaga, Jiaxuan You, Matei Zaharia, Michael Zhang, Tianyi Zhang, Xikun Zhang, Yuhui Zhang, Lucia Zheng, Kaitlyn Zhou, and Percy Liang.
\newblock On the opportunities and risks of foundation models, July 2022.
\newblock arXiv:2108.07258.

\bibitem{Bert2019}
Jacob Devlin, Ming-Wei Chang, Kenton Lee, and Kristina Toutanova.
\newblock {BERT}: {Pre}-training of deep bidirectional transformers for language understanding.
\newblock In {\em Proc. {NAACL}-{HLT} 2019}, pages 4171--4186, June 2019.

\bibitem{zhangAnomalyDetectionControl2023}
Kuan Zhang, Shuchen Wang, Saijin Wang, and Qizhi Xu.
\newblock Anomaly detection of control moment gyroscope based on working condition classification and transfer learning.
\newblock {\em Applied Sciences}, 13(7):4259, January 2023.

\bibitem{vaswani2017attention}
Ashish Vaswani, Noam Shazeer, Niki Parmar, Jakob Uszkoreit, Llion Jones, Aidan~N Gomez, Lukasz Kaiser, and Illia Polosukhin.
\newblock Attention is all you need.
\newblock In {\em Proc. {NeurIPS}}, volume~30, 2017.

\bibitem{dou2021gpt}
Yao Dou, Maxwell Forbes, Rik Koncel-Kedziorski, Noah~A. Smith, and Yejin Choi.
\newblock Is {GPT}-3 text indistinguishable from human text? {Scarecrow}: {A} framework for scrutinizing machine text.
\newblock In {\em Proc. 60th {Annu}. {Meeting}. {ACL}}, pages 7250--7274, May 2022.

\bibitem{alexandr2021fine}
Nikolich Alexandr, Osliakova Irina, Kudinova Tatyana, Kappusheva Inessa, and Puchkova Arina.
\newblock Fine-tuning {GPT}-3 for {Russian} text summarization.
\newblock In {\em Proc. {Data} {Science} and {Intelligent} {Systems}}, pages 748--757, 2021.

\bibitem{Guo2019SpotTune}
Yunhui Guo, Honghui Shi, Abhishek Kumar, Kristen Grauman, Tajana Rosing, and Rogerio Feris.
\newblock {SpotTune}: {Transfer} learning through adaptive fine-tuning.
\newblock In {\em Proc. {CVPR}}, pages 4800--4809, June 2019.

\bibitem{sager2022unsupervised}
Pascal Sager, Sebastian Salzmann, Felice Burn, and Thilo Stadelmann.
\newblock Unsupervised domain adaptation for vertebrae detection and identification in {3D} {CT} volumes using a domain sanity loss.
\newblock {\em J. Imaging}, 8(8), 2022.

\bibitem{NEURIPS2020_1457c0d6}
Tom Brown, Benjamin Mann, Nick Ryder, Melanie Subbiah, Jared~D Kaplan, Prafulla Dhariwal, Arvind Neelakantan, Pranav Shyam, Girish Sastry, Amanda Askell, Sandhini Agarwal, Ariel Herbert-Voss, Gretchen Krueger, Tom Henighan, Rewon Child, Aditya Ramesh, Daniel Ziegler, Jeffrey Wu, Clemens Winter, Chris Hesse, Mark Chen, Eric Sigler, Mateusz Litwin, Scott Gray, Benjamin Chess, Jack Clark, Christopher Berner, Sam McCandlish, Alec Radford, Ilya Sutskever, and Dario Amodei.
\newblock Language models are few-shot learners.
\newblock In {\em Proc. {NeurIPS}}, volume~33, pages 1877--1901, 2020.

\bibitem{wangFewShotTransferLearning2022}
Yanxin Wang, Jing Yan, Xinyu Ye, Qianzhen Jing, Jianhua Wang, and Yingsan Geng.
\newblock Few-shot transfer learning with attention mechanism for high-voltage rircuit breaker fault diagnosis.
\newblock {\em IEEE Trans. Ind. Appl.}, 58(3):3353--3360, May 2022.

\bibitem{tzeng2014deep}
Eric Tzeng, Judy Hoffman, Ning Zhang, Kate Saenko, and Trevor Darrell.
\newblock Deep domain confusion: {Maximizing} for domain invariance, December 2014.
\newblock arXiv:1412.3474.

\bibitem{long2017deep}
Mingsheng Long, Han Zhu, Jianmin Wang, and Michael~I Jordan.
\newblock Deep transfer learning with joint adaptation networks.
\newblock In {\em Proc. {ICML}}, pages 2208--2217, July 2017.

\bibitem{long2015learning}
Mingsheng Long, Yue Cao, Jianmin Wang, and Michael Jordan.
\newblock Learning transferable features with deep adaptation networks.
\newblock In {\em Proc. {ICML}}, volume~37, pages 97--105, July 2015.

\bibitem{zhang2015deep}
Xu~Zhang, Felix~Xinnan Yu, Shih-Fu Chang, and Shengjin Wang.
\newblock Deep transfer network: {Unsupervised} domain adaptation, March 2015.
\newblock arXiv:1503.00591.

\bibitem{venkateswara2017deep}
Hemanth Venkateswara, Jose Eusebio, Shayok Chakraborty, and Sethuraman Panchanathan.
\newblock Deep hashing network for unsupervised domain adaptation.
\newblock In {\em Proc. {CVPR}}, pages 5385--5394, July 2017.

\bibitem{soleimaniCrosssubjectTransferLearning2021}
Elnaz Soleimani and Ehsan Nazerfard.
\newblock Cross-subject transfer learning in human activity recognition systems using generative adversarial networks.
\newblock {\em Neurocomputing}, 426:26--34, February 2021.

\bibitem{tzeng2015simultaneous}
Eric Tzeng, Judy Hoffman, Trevor Darrell, and Kate Saenko.
\newblock Simultaneous deep transfer across domains and tasks.
\newblock In {\em Proc. {ICCV}}, pages 4068--4076, December 2015.

\bibitem{ozyurtContrastiveLearningUnsupervised2023}
Yilmazcan Ozyurt, Stefan Feuerriegel, and Ce~Zhang.
\newblock Contrastive learning for unsupervised domain adaptation of time series, February 2023.
\newblock arXiv:2206.06243.

\bibitem{ganin2016domain}
Yaroslav Ganin, Evgeniya Ustinova, Hana Ajakan, Pascal Germain, Hugo Larochelle, François Laviolette, Mario Marchand, and Victor Lempitsky.
\newblock Domain-adversarial training of neural networks.
\newblock In {\em Domain {Adaptation} in {Computer} {Vision} {Applications}}, pages 189--209. Springer, 2017.

\bibitem{ajakan2014domain}
Hana Ajakan, Pascal Germain, Hugo Larochelle, François Laviolette, and Mario Marchand.
\newblock Domain-adversarial neural networks, February 2015.
\newblock arXiv:1412.4446.

\bibitem{Tzeng_2017_CVPR}
Eric Tzeng, Judy Hoffman, Kate Saenko, and Trevor Darrell.
\newblock Adversarial discriminative domain adaptation.
\newblock In {\em Proc. {CVPR}}, pages 7167--7176, July 2017.

\bibitem{sun2021multilingual}
Zewei Sun, Mingxuan Wang, and Lei Li.
\newblock Multilingual translation via grafting pre-trained language models.
\newblock In {\em Proc. {EMNLP} 2021}, pages 2735--2747, November 2021.

\bibitem{glass2019span}
Michael Glass, Alfio Gliozzo, Rishav Chakravarti, Anthony Ferritto, Lin Pan, G~P~Shrivatsa Bhargav, Dinesh Garg, and Avi Sil.
\newblock Span selection pre-training for question answering.
\newblock In {\em Proc. 58th {Annu}. {Meeting}. {ACL}}, pages 2773--2782, July 2020.

\bibitem{tuggener2021imagenet}
Lukas Tuggener, Jürgen Schmidhuber, and Thilo Stadelmann.
\newblock Is it enough to optimize {CNN} architectures on {ImageNet}?
\newblock {\em Frontiers in Computer Science}, 4, 2022.

\bibitem{shenWassersteinDistanceGuided2018}
Jian Shen, Yanru Qu, Weinan Zhang, and Yong Yu.
\newblock Wasserstein distance guided representation learning for domain adaptation.
\newblock In {\em Proc. {AAAI}}, pages 4058--4065, February 2018.

\bibitem{daiCoclusteringBasedClassification2007}
Wenyuan Dai, Gui-Rong Xue, Qiang Yang, and Yong Yu.
\newblock Co-clustering based classification for out-of-domain documents.
\newblock In {\em Proc. 13th {ACM} {SIGKDD}}, {KDD} '07, pages 210--219, August 2007.

\bibitem{chatfieldReturnDevilDetails2014}
Ken Chatfield, Karen Simonyan, Andrea Vedaldi, and Andrew Zisserman.
\newblock Return of the devil in the details: {Delving} deep into convolutional nets.
\newblock In {\em Proc. {BMVC} 2014}, pages 6.1--6.12, 2014.

\bibitem{goodfellow2020generative}
Ian Goodfellow, Jean Pouget-Abadie, Mehdi Mirza, Bing Xu, David Warde-Farley, Sherjil Ozair, Aaron Courville, and Yoshua Bengio.
\newblock Generative adversarial nets.
\newblock In {\em Proc. {NeurIPS}}, volume~27, 2014.

\bibitem{SCHMIDHUBER202058}
Jürgen Schmidhuber.
\newblock Generative {Adversarial} {Networks} are special cases of {Artificial} {Curiosity} (1990) and also closely related to {Predictability} {Minimization} (1991).
\newblock {\em Neural Networks}, 127:58--66, 2020.

\bibitem{parisi2019continual}
German~I. Parisi, Ronald Kemker, Jose~L. Part, Christopher Kanan, and Stefan Wermter.
\newblock Continual lifelong learning with neural networks: {A} review.
\newblock {\em Neural Networks}, 113:54--71, May 2019.

\bibitem{wang2020generalizing}
Yaqing Wang, Quanming Yao, James~T Kwok, and Lionel~M Ni.
\newblock Generalizing from a few examples: {A} survey on few-shot learning.
\newblock {\em ACM Comput. Surv.}, 53(3):63:1--63:34, June 2020.

\bibitem{fei2006one}
Li~Fei-Fei, R.~Fergus, and P.~Perona.
\newblock One-shot learning of object categories.
\newblock {\em IEEE TPAMI}, 28(4):594--611, April 2006.

\bibitem{lampert2009learning}
Christoph~H Lampert, Hannes Nickisch, and Stefan Harmeling.
\newblock Learning to detect unseen object classes by between-class attribute transfer.
\newblock In {\em Proc. {CVPR}}, pages 951--958, June 2009.

\bibitem{zhouDomainGeneralizationSurvey2023}
Kaiyang Zhou, Ziwei Liu, Yu~Qiao, Tao Xiang, and Chen~Change Loy.
\newblock Domain generalization: {A} survey.
\newblock {\em IEEE TPAMI}, 45(4):4396--4415, April 2023.

\bibitem{blanchardGeneralizingSeveralRelated2011}
Gilles Blanchard, Gyemin Lee, and Clayton Scott.
\newblock Generalizing from several related classification tasks to a new unlabeled sample.
\newblock In {\em Proc. {NeurIPS}}, volume~24, 2011.

\bibitem{finn2017model}
Chelsea Finn, Pieter Abbeel, and Sergey Levine.
\newblock Model-agnostic meta-learning for fast adaptation of deep networks.
\newblock In {\em Proc. 34th {ICML}}, pages 1126--1135. PMLR, July 2017.

\bibitem{hospedales2021meta}
Timothy Hospedales, Antreas Antoniou, Paul Micaelli, and Amos Storkey.
\newblock Meta-learning in neural networks: {A} survey.
\newblock {\em IEEE TPAMI}, 44(09):5149--5169, September 2022.

\bibitem{gouKnowledgeDistillationSurvey2021}
Jianping Gou, Baosheng Yu, Stephen~J. Maybank, and Dacheng Tao.
\newblock Knowledge distillation: {A} survey.
\newblock {\em Int. J. Comput. Vision}, 129(6):1789--1819, June 2021.

\bibitem{liuSelfSupervisedLearningGenerative2023}
Xiao Liu, Fanjin Zhang, Zhenyu Hou, Li~Mian, Zhaoyu Wang, Jing Zhang, and Jie Tang.
\newblock Self-supervised learning: {Generative} or contrastive.
\newblock {\em IEEE Trans. Knowl. Data Eng.}, 35(1):857--876, January 2023.

\bibitem{bai2021self}
Haoping Bai, Meng Cao, Ping Huang, and Jiulong Shan.
\newblock Self-supervised semi-supervised learning for data labeling and quality evaluation.
\newblock In {\em {NeuIPS}. {Workshop}}, November 2021.

\bibitem{Tercan2018bridging}
Hasan Tercan, Alexandro Guajardo, Julian Heinisch, Thomas Thiele, Christian Hopmann, and Tobias Meisen.
\newblock Transfer-learning: {Bridging} the gap between real and simulation data for machine learning in injection molding.
\newblock {\em Procedia CIRP}, 72:185--190, January 2018.

\bibitem{gornitz2013toward}
Nico Görnitz, Marius Kloft, Konrad Rieck, and Ulf Brefeld.
\newblock Toward supervised anomaly detection.
\newblock {\em JAIR}, 46:235--262, February 2013.

\bibitem{zhang2021unsupervised}
Yuxin Zhang, Yiqiang Chen, Jindong Wang, and Zhiwen Pan.
\newblock Unsupervised deep anomaly detection for multi-sensor time-series signals.
\newblock {\em IEEE Trans. Knowl. Data Eng.}, 35(2):2118--2132, February 2023.

\bibitem{stadelmann2019beyond}
Thilo Stadelmann, Vasily Tolkachev, Beate Sick, Jan Stampfli, and Oliver Dürr.
\newblock Beyond {ImageNet}: {Deep} learning in industrial practice.
\newblock In {\em Applied {Data} {Science}: {Lessons} {Learned} for the {Data}-{Driven} {Business}}, pages 205--232. Springer, 2019.

\bibitem{hawkins1980identification}
Douglas~M Hawkins.
\newblock {\em Identification of outliers}, volume~11.
\newblock Springer Netherlands, Dordrecht, 1980.

\bibitem{zhuInterpretableMultivariateTimeseries2023}
Haiqi Zhu, Chunzhi Yi, Seungmin Rho, Shaohui Liu, and Feng Jiang.
\newblock An interpretable multivariate time-series anomaly detection method in cyber-physical systems based on adaptive mask.
\newblock {\em IEEE Internet Things J.}, pages 1--1, 2023.

\bibitem{schmidlAnomalyDetectionTime2022a}
Sebastian Schmidl, Phillip Wenig, and Thorsten Papenbrock.
\newblock Anomaly detection in time series: {A} comprehensive evaluation.
\newblock {\em Proc. VLDB Endow.}, 15(9):1779--1797, May 2022.

\bibitem{audibert2020usad}
Julien Audibert, Pietro Michiardi, Frédéric Guyard, Sébastien Marti, and Maria~A. Zuluaga.
\newblock {USAD}: {Unsupervised} anomaly detection on multivariate time series.
\newblock In {\em Proc. 26th {ACM} {SIGKDD}}, {KDD} '20, pages 3395--3404, 2020.

\bibitem{malhotraLSTMbasedEncoderDecoderMultisensor2016b}
Pankaj Malhotra, Anusha Ramakrishnan, Gaurangi Anand, Lovekesh Vig, Puneet Agarwal, and Gautam Shroff.
\newblock {LSTM}-based encoder-decoder for multi-sensor anomaly detection, July 2016.
\newblock arXiv:1607.00148.

\bibitem{weiLSTMAutoencoderBasedAnomalyDetection2023}
Yuanyuan Wei, Julian Jang-Jaccard, Wen Xu, Fariza Sabrina, Seyit Camtepe, and Mikael Boulic.
\newblock {LSTM}-autoencoder-based anomaly detection for indoor air quality time-series data.
\newblock {\em IEEE Sens. J.}, 23(4):3787--3800, February 2023.

\bibitem{zengMultivariateTimeSeries2023}
Fanyu Zeng, Mengdong Chen, Cheng Qian, Yanyang Wang, Yijun Zhou, and Wenzhong Tang.
\newblock Multivariate time series anomaly detection with adversarial transformer architecture in the {Internet} of {Things}.
\newblock {\em Future Gener. Comput. Syst.}, 144:244--255, July 2023.

\bibitem{liMADGANMultivariateAnomaly2019}
Dan Li, Dacheng Chen, Baihong Jin, Lei Shi, Jonathan Goh, and See-Kiong Ng.
\newblock {MAD}-{GAN}: {Multivariate} anomaly detection for time series data with generative adversarial networks.
\newblock In {\em Proc. {ICANN} 2019}, pages 703--716, 2019.

\bibitem{niuVAEGANTimeSeries2020}
Zijian Niu, Ke~Yu, and Xiaofei Wu.
\newblock {LSTM}-based {VAE}-{GAN} for time-series anomaly detection.
\newblock {\em Sensors}, 20(13):3738, January 2020.

\bibitem{basharTAnoGANTimeSeries2020}
Md~Abul Bashar and Richi Nayak.
\newblock {TAnoGAN}: {Time} series anomaly detection with generative adversarial networks.
\newblock In {\em Proc. {IEEE} {SSCI}}, pages 1778--1785, December 2020.

\bibitem{kimTimeseriesAnomalyDetection2023}
Jina Kim, Hyeongwon Kang, and Pilsung Kang.
\newblock Time-series anomaly detection with stacked {Transformer} representations and {1D} convolutional network.
\newblock {\em EAAI}, 120:105964, April 2023.

\bibitem{dengGraphNeuralNetworkBased2021}
Ailin Deng and Bryan Hooi.
\newblock Graph neural network-based anomaly detection in multivariate time series.
\newblock {\em Proc. AAAI}, 35(5):4027--4035, May 2021.

\bibitem{tangGRUBasedInterpretableMultivariate2023}
Chaofan Tang, Lijuan Xu, Bo~Yang, Yongwei Tang, and Dawei Zhao.
\newblock {GRU}-based interpretable multivariate time series anomaly detection in industrial control system.
\newblock {\em Comput. Secur.}, 127:103094, April 2023.

\bibitem{dingMSTGATMultimodalSpatial2023}
Chaoyue Ding, Shiliang Sun, and Jing Zhao.
\newblock {MST}-{GAT}: {A} multimodal spatial–temporal graph attention network for time series anomaly detection.
\newblock {\em Inf. Fusion}, 89:527--536, January 2023.

\bibitem{himeurNovelApproachDetecting2020}
Yassine Himeur, Abdullah Alsalemi, Faycal Bensaali, and Abbes Amira.
\newblock A novel approach for detecting anomalous energy consumption based on micro-moments and deep neural networks.
\newblock {\em Cognit. Comput.}, 12(6):1381--1401, November 2020.

\bibitem{yangDCdetectorDualAttention2023}
Yiyuan Yang, Chaoli Zhang, Tian Zhou, Qingsong Wen, and Liang Sun.
\newblock {DCdetector}: {Dual} attention contrastive representation learning for time series anomaly detection.
\newblock In {\em Proc. 29th {ACM} {SIGKDD}}, {KDD} '23, pages 3033--3045, August 2023.

\bibitem{li2022perspective}
Weihua Li, Ruyi Huang, Jipu Li, Yixiao Liao, Zhuyun Chen, Guolin He, Ruqiang Yan, and Konstantinos Gryllias.
\newblock A perspective survey on deep transfer learning for fault diagnosis in industrial scenarios: {Theories}, applications and challenges.
\newblock {\em Mech. Syst. Signal Process.}, 167:108487, March 2022.

\bibitem{ma2019improving}
Jun Ma, Jack~CP Cheng, Changqing Lin, Yi~Tan, and Jingcheng Zhang.
\newblock Improving air quality prediction accuracy at larger temporal resolutions using deep learning and transfer learning techniques.
\newblock {\em Atmos. Environ.}, 214:116885, October 2019.

\bibitem{rosenberg2018end}
Ishai Rosenberg, Guillaume Sicard, and Eli David.
\newblock End-to-end deep neural networks and transfer learning for automatic analysis of nation-state malware.
\newblock {\em Entropy}, 20(5):390, May 2018.

\bibitem{pan2023transfer}
Qiuyue Pan, Yuequan Bao, and Hui Li.
\newblock Transfer learning-based data anomaly detection for structural health monitoring.
\newblock {\em Struct. Health Monit.}, 22(5):3077--3091, January 2023.

\bibitem{ComparativeStudyDeep}
Guannan Li, Liang Chen, Jiangyan Liu, and Xi~Fang.
\newblock Comparative study on deep transfer learning strategies for cross-system and cross-operation-condition building energy systems fault diagnosis.
\newblock {\em Energy}, 263:125943, January 2023.

\bibitem{serradilla2021adaptable}
Oscar Serradilla, Ekhi Zugasti, Julian Ramirez~de Okariz, Jon Rodriguez, and Urko Zurutuza.
\newblock Adaptable and explainable predictive maintenance: {Semi}-supervised deep learning for anomaly detection and diagnosis in press machine data.
\newblock {\em Applied Sciences}, 11(16):7376, January 2021.

\bibitem{zgraggen2021transfer}
Jannik Zgraggen, Markus Ulmer, Eskil Jarlskog, Gianmarco Pizza, and Lilach~Goren Huber.
\newblock Transfer learning approaches for wind turbine fault detection using deep learning.
\newblock {\em PHME 2021}, 6(1):12--12, June 2021.

\bibitem{zabin2022hybrid}
Mahe Zabin, Ho-Jin Choi, and Jia Uddin.
\newblock Hybrid deep transfer learning architecture for industrial fault diagnosis using {Hilbert} transform and {DCNN}–{LSTM}.
\newblock {\em J. Supercomput.}, 79(5):5181--5200, March 2023.

\bibitem{tercan2019industrial}
Hasan Tercan, Alexandro Guajardo, and Tobias Meisen.
\newblock Industrial transfer learning: {Boosting} machine learning in production.
\newblock In {\em Proc. {IEEE} 17th {INDIN}}, volume~1, pages 274--279, July 2019.

\bibitem{lockner2022transfer}
Yannik Lockner, Christian Hopmann, and Weibo Zhao.
\newblock Transfer learning with artificial neural networks between injection molding processes and different polymer materials.
\newblock {\em J. Manuf. Processes}, 73:395--408, January 2022.

\bibitem{gellrichDeepTransferLearning2021}
Sebastian Gellrich, Marc-André Filz, Anna-Sophia Wilde, Thomas Beganovic, Alexander Mattheus, Tim Abraham, and Christoph Herrmann.
\newblock Deep transfer learning for improved product quality prediction: {A} case study of {Aluminum} gravity die casting.
\newblock {\em Procedia CIRP}, 104:912--917, January 2021.

\bibitem{abdallah2023anomaly}
Mustafa Abdallah, Byung-Gun Joung, Wo~Jae Lee, Charilaos Mousoulis, Nithin Raghunathan, Ali Shakouri, John~W. Sutherland, and Saurabh Bagchi.
\newblock Anomaly detection and inter-sensor transfer learning on smart manufacturing datasets.
\newblock {\em Sensors}, 23(1):486, January 2023.

\bibitem{maschler2021regularization}
Benjamin Maschler, Thi~Thu Huong~Pham, and Michael Weyrich.
\newblock Regularization-based continual learning for anomaly detection in discrete manufacturing.
\newblock {\em Procedia CIRP}, 104:452--457, January 2021.

\bibitem{parkMENDELTimeSeries}
Jeongyong Park, Bedeuro Kim, and Hyoungshick Kim.
\newblock {MENDEL}: {Time} series anomaly detection using transfer learning for industrial control systems.
\newblock In {\em Proc. {IEEE} {BigComp}}, pages 261--268, February 2023.

\bibitem{liang2018transfer}
Peng Liang, Hai-Dong Yang, Wen-Si Chen, Si-Yuan Xiao, and Zhao-Ze Lan.
\newblock Transfer learning for aluminium extrusion electricity consumption anomaly detection via deep neural networks.
\newblock {\em Int. J. Comput. Integr. Manuf.}, 31(4-5):396--405, April 2018.

\bibitem{copiaco2023innovative}
Abigail Copiaco, Yassine Himeur, Abbes Amira, Wathiq Mansoor, Fodil Fadli, Shadi Atalla, and Shahab~Saquib Sohail.
\newblock An innovative deep anomaly detection of building energy consumption using energy time-series images.
\newblock {\em EAAI}, 119:105775, March 2023.

\bibitem{xu2021anomaly}
Chuqiao Xu, Junliang Wang, Jie Zhang, and Xiaoou Li.
\newblock Anomaly detection of power consumption in yarn spinning using transfer learning.
\newblock {\em Comput. Ind. Eng.}, 152:107015, February 2021.

\bibitem{simoneAnalysisMachineLearninga}
Francesco~Di Simone and Francesco Amigoni.
\newblock Analysis of machine learning methods for anomaly detection of power consumption in buildings.
\newblock Master's thesis, Politecnico di Milano, 2021.
\newblock Available: https://www.politesi.polimi.it/handle/10589/183344.

\bibitem{hsieh2019unsupervised}
Ruei-Jie Hsieh, Jerry Chou, and Chih-Hsiang Ho.
\newblock Unsupervised online anomaly detection on multivariate sensing time series data for smart manufacturing.
\newblock In {\em Proc. {IEEE} 12th {SOCA}}, pages 90--97, November 2019.

\bibitem{serradilla2022deep}
Oscar Serradilla, Ekhi Zugasti, Jon Rodriguez, and Urko Zurutuza.
\newblock Deep learning models for predictive maintenance: a survey, comparison, challenges and prospects.
\newblock {\em Appl. Intell.}, 52(10):10934--10964, August 2022.

\bibitem{wen2017new}
Long Wen, Liang Gao, and Xinyu Li.
\newblock A new deep transfer learning based on sparse auto-encoder for fault diagnosis.
\newblock {\em IEEE Trans. Syst. Man Cybern. Syst.}, 49(1):136--144, January 2019.

\bibitem{Ronneberger2015UNet}
Olaf Ronneberger, Philipp Fischer, and Thomas Brox.
\newblock U-net: {Convolutional} networks for biomedical image segmentation.
\newblock In {\em Proc. {MICCAI}}, pages 234--241, 2015.

\bibitem{8540073}
Chuang Sun and {others}.
\newblock Deep transfer learning based on sparse autoencoder for remaining useful life prediction of tool in manufacturing.
\newblock {\em IEEE Trans. Ind. Inf.}, 15(4):2416--2425, April 2019.

\bibitem{copiacoExploringDeepTimeSeries2022}
Abigail Copiaco, Yassine Himeur, Abbes Amira, Wathiq Mansoor, Fodil Fadli, and Shadi Atalla.
\newblock Exploring deep time-series imaging for anomaly detection of building energy consumption.
\newblock In {\em Proc. {IEEE} {CSDE}}, pages 1--5, December 2022.

\bibitem{sayedDeepTransferLearning2022}
Aya~Nabil Sayed, Yassine Himeur, and Faycal Bensaali.
\newblock Deep and transfer learning for building occupancy detection: {A} review and comparative analysis.
\newblock {\em EAAI}, 115:105254, October 2022.

\bibitem{goldstein2016comparative}
Markus Goldstein and Seiichi Uchida.
\newblock A comparative evaluation of unsupervised anomaly detection algorithms for multivariate data.
\newblock {\em PloS one}, 11(4):1--31, April 2016.

\bibitem{buda2018systematic}
Mateusz Buda, Atsuto Maki, and Maciej~A. Mazurowski.
\newblock A systematic study of the class imbalance problem in convolutional neural networks.
\newblock {\em Neural Networks}, 106:249--259, October 2018.

\bibitem{smith2001transfer}
Kimberly~A Smith-Jentsch, Eduardo Salas, and Michael~T Brannick.
\newblock To transfer or not to transfer? {Investigating} the combined effects of trainee characteristics, team leader support, and team climate.
\newblock {\em J. Appl. Psychol}, 86:279--92, May 2001.

\bibitem{glorotDomainAdaptationLargeScale2011}
Xavier Glorot, Antoine Bordes, and Yoshua Bengio.
\newblock Domain adaptation for large-scale sentiment classification: a deep learning approach.
\newblock In {\em Proc. 28th {ICML}}, pages 513--520, June 2011.

\bibitem{8853246}
Wenqian Jiang, Yang Hong, Beitong Zhou, Xin He, and Cheng Cheng.
\newblock A {GAN}-based anomaly detection approach for imbalanced industrial time series.
\newblock {\em IEEE Access}, 7:143608--143619, 2019.

\bibitem{goodfellow2014explaining}
Ian~J. Goodfellow, Jonathon Shlens, and Christian Szegedy.
\newblock Explaining and harnessing adversarial examples.
\newblock In {\em Proc. {ICLR}}, 2015.

\bibitem{chawla2002smote}
N.~V. Chawla, K.~W. Bowyer, L.~O. Hall, and W.~P. Kegelmeyer.
\newblock {SMOTE}: {Synthetic} minority over-sampling technique.
\newblock {\em JAIR}, 16:321--357, June 2002.

\bibitem{ijaz2018hybrid}
Muhammad~Fazal Ijaz, Ganjar Alfian, Muhammad Syafrudin, and Jongtae Rhee.
\newblock Hybrid prediction model for type 2 diabetes and hypertension using {DBSCAN}-based outlier detection, synthetic minority over sampling technique ({SMOTE}), and random forest.
\newblock {\em Applied Sciences}, 8(8):1325, August 2018.

\bibitem{mokhtari2021machine}
Sohrab Mokhtari, Alireza Abbaspour, Kang~K Yen, and Arman Sargolzaei.
\newblock A machine learning approach for anomaly detection in industrial control systems based on measurement data.
\newblock {\em Electronics}, 10(4):407, January 2021.

\bibitem{segmentationAnything2023}
Alexander Kirillov, Eric Mintun, Nikhila Ravi, Hanzi Mao, Chloe Rolland, Laura Gustafson, Tete Xiao, Spencer Whitehead, Alexander~C. Berg, Wan-Yen Lo, Piotr Dollar, and Ross Girshick.
\newblock Segment anything.
\newblock In {\em Proc. {ICCV}}, pages 4015--4026, 2023.

\bibitem{Rombach2022}
Robin Rombach, Andreas Blattmann, Dominik Lorenz, Patrick Esser, and Björn Ommer.
\newblock High-resolution image synthesis with latent diffusion models.
\newblock In {\em Proc. {CVPR}}, pages 10684--10695, June 2022.

\bibitem{estimating2023}
Athanasios Vlontzos, Bernhard Kainz, and Ciarán~M. Gilligan-Lee.
\newblock Estimating categorical counterfactuals via deep twin networks.
\newblock {\em Nat. Mach. Intell.}, 5(2):159--168, February 2023.

\bibitem{wang2018edge}
Shiqiang Wang, Tiffany Tuor, Theodoros Salonidis, Kin~K. Leung, Christian Makaya, Ting He, and Kevin Chan.
\newblock When edge meets learning: {Adaptive} control for resource-constrained distributed machine learning.
\newblock In {\em Proc. {IEEE} {INFOCOM} 2018}, pages 63--71, April 2018.

\bibitem{amiri2020machine}
Mohammad~Mohammadi Amiri and Deniz Gündüz.
\newblock Machine learning at the wireless edge: {Distributed} stochastic gradient descent over-the-air.
\newblock In {\em Proc. {IEEE} {ISIT}}, pages 1432--1436, July 2019.

\bibitem{Ge_2017_CVPR}
Weifeng Ge and Yizhou Yu.
\newblock Borrowing treasures from the wealthy: {Deep} transfer learning through selective joint fine-tuning.
\newblock In {\em Proc. {IEEE} {CVPR}}, pages 10--19, July 2017.

\bibitem{xuAnomalyTransformerTime2021}
Jiehui Xu, Haixu Wu, Jianmin Wang, and Mingsheng Long.
\newblock Anomaly {Transformer}: {Time} series anomaly detection with association discrepancy.
\newblock In {\em {ICLR}}, October 2021.

\bibitem{pintilie2023}
Ioana Pintilie, Andrei Manolache, and Florin Brad.
\newblock Time series anomaly detection using diffusion-based models, November 2023.
\newblock arXiv:2311.01452.

\bibitem{stadelmann2022data}
Thilo Stadelmann, Tino Klamt, and Philipp~H Merkt.
\newblock Data centrism and the core of {Data} {Science} as a scientific discipline.
\newblock {\em Archives of Data Science, Series A}, 8(2), March 2022.

\bibitem{luley2023}
Paul-Philipp Luley, Jan~Milan Deriu, Peng Yan, Gerrit~A. Schatte, and Thilo Stadelmann.
\newblock From concept to implementation: the data-centric development process for {AI} in industry.
\newblock In {\em Proc. 10th {IEEE} {SDS}}, 2023.

\bibitem{liu2021autodc}
Zac Yung-Chun Liu, Shoumik Roychowdhury, Scott Tarlow, Akash Nair, Shweta Badhe, and Tejas Shah.
\newblock {AutoDC}: {Automated} data-centric processing.
\newblock In {\em {NeuIPS}. {Workshop}}, November 2021.
\newblock Available: https://nips.cc/virtual/2021/38244.

\bibitem{kang2022finding}
Daniel Kang, Nikos Arechiga, Sudeep Pillai, Peter~D. Bailis, and Matei Zaharia.
\newblock Finding label and model errors in perception data with learned observation assertions.
\newblock In {\em Proc. {ACM} {SIGMOD}}, pages 496--505, June 2022.

\bibitem{salemAnomalyGenerationUsing2018}
Milad Salem, Shayan Taheri, and Jiann~Shiun Yuan.
\newblock Anomaly generation using generative adversarial networks in host-based intrusion detection.
\newblock In {\em Proc. {IEEE} {UEMCON}}, pages 683--687, November 2018.

\end{thebibliography}
\end{document}